\documentclass{article}
\usepackage{ijcai24}

\usepackage{times}
\usepackage{soul}
\usepackage{url}
\usepackage[hidelinks]{hyperref}
\usepackage[utf8]{inputenc}
\usepackage[small]{caption}
\usepackage{graphicx}
\usepackage{amsmath}
\usepackage{amsthm}
\usepackage{booktabs}
\usepackage{algorithm}
\usepackage{algorithmic} 
\usepackage[switch]{lineno}
\usepackage[toc,page]{appendix}

\title{Learning Logic Programs by Finding Minimal Unsatisfiable Subprograms}

\author{
Andrew Cropper
\and
C\'{e}line Hocquette
\affiliations
University of Oxford\\
\emails
\{andrew.cropper,celine.hocquette\}@cs.ox.ac.uk
}

\usepackage[utf8]{inputenc}
\usepackage{xcolor}
\usepackage{inconsolata}
\usepackage{tikz}
\usepackage{pgfplots}
\usepackage{microtype}
\usepackage{amssymb}
\usepackage{amsthm}
\usepackage{algorithm}
\usepackage{subfig}

\usepackage{booktabs}

\theoremstyle{definition}
\newtheorem{definition}{Definition}
\newtheorem{example}{Example}
\newtheorem{theorem}{Theorem}
\newtheorem{proposition}{Proposition}

\newcommand{\name}{\textsc{MUSPer}}
\newcommand{\popper}{\textsc{Popper}}
\newcommand{\ale}{\textsc{aleph}}
\newcommand{\ilasp}{\textsc{ilasp}}
\newcommand{\aspal}{\textsc{aspal}}
\newcommand{\metagol}{\textsc{metagol}}
\newcommand{\disco}{\textsc{Disco}}

\usepackage{listings}

\lstnewenvironment{myalgorithm}[1][] 
{
    \lstset{ 
        basicstyle=\ttfamily\small,
        columns=flexible,
        showspaces=false,               
        showstringspaces=false,         
        mathescape=true,
        numbers=left,
        escapeinside={*}{*},
        columns=flexible,
        escapechar=@,
        numbersep=3pt,        
        keywordstyle=\bfseries,
        keywords={,and, return, not, def, in,elif,if, else, for, foreach, while,}
        numbers=left,
        xleftmargin=8pt,
        #1 
    }
}
{}

\begin{document}

\maketitle

\begin{abstract}
    The goal of inductive logic programming (ILP) is to search for a logic program that generalises training examples and background knowledge.    
    We introduce an ILP approach that identifies \emph{minimal unsatisfiable subprograms} (MUSPs).
    We show that finding MUSPs allows us to efficiently and soundly prune the search space.    
    Our experiments on multiple domains, including program synthesis and game playing, show that our approach can reduce learning times by 99\%.
\end{abstract}
\section{Introduction}

The goal of inductive logic programming (ILP) is to induce a logic program (a set of logical rules) that generalises examples and background knowledge (BK).
For instance, suppose we have BK with the relations \emph{head/2}, \emph{tail/2}, and \emph{empty/1} and the following positive ($E^+$) and negative ($E^-$) examples:

\begin{center}
\begin{tabular}{l}
\emph{E$^+$ = \{f([i,j,c,a,i],i), f([a,a,a,i],i), f([i,l,p],p)\}}\\
\emph{E$^-$ = \{f([i,j,c,a,i],j), f([a,a,a,i],a), f([i,l,p],i)\}}\\
\end{tabular}
\end{center}

\noindent
Given these inputs, we might want to learn a program that computes the \emph{last} element of a list, such as:
\[
    \begin{array}{l}
    \left\{
    \begin{array}{l}
        \emph{f(A,B) $\leftarrow$ tail(A,C), empty(C), head(A,B)}\\    
        \emph{f(A,B) $\leftarrow$ tail(A,C), f(C,B)}
    \end{array}
    \right\}
    \end{array}
\]

\noindent
To find a program, a learner (an ILP system) searches a hypothesis space (the set of all programs) for a \emph{solution} (a program that correctly generalises the examples).
A learner tests programs on the examples and uses the outcome to guide the search.
For instance, suppose a learner tests the program:
\[
    \begin{array}{l}
    \left\{
    \begin{array}{l}
        \emph{f(A,B) $\leftarrow$ head(A,B)}    
    \end{array}
    \right\}
    \end{array}
\]
\noindent
This program is too general as it entails the negative example \emph{f([i,l,p],i)}.
A learner can, therefore, prune more general programs, such as:

\[
    \begin{array}{l}
    \left\{
    \begin{array}{l}
        \emph{f(A,B) $\leftarrow$ head(A,B)}\\    
        \emph{f(A,B) $\leftarrow$ tail(A,C), head(C,B)}
    \end{array}
    \right\}
    \end{array}
\]
\noindent
Similarly, suppose a learner tests the program:
\[
    \begin{array}{l}
    h = \left\{
    \begin{array}{l}
        \emph{f(A,B) $\leftarrow$ empty(A), head(A,B), tail(A,C), head(C,B)}
    \end{array}
    \right\}
    \end{array}
\]
\noindent
The program $h$ is too specific because it does not entail any positive examples.
A learner can, therefore, prune more specific programs.
However, the simple explanation that $h$ is too specific is rather crude.
The program $h$ cannot entail \emph{anything} because an empty list cannot have a tail.
In other words, the pair of literals \emph{empty(A)} and \emph{tail(A,C)} is unsatisfiable.
Therefore, any rule that contains this pair is also unsatisfiable, so we can prune all such rules from the hypothesis space.

We argue that by finding minimal unsatisfiable programs we can prune more programs and thus improve learning performance.
This idea is inspired by work in SAT on finding \emph{minimal unsatisfiable cores}  \cite{unsat_core}.
We therefore use similar terminology and say that the pair of literals \emph{empty(A)} and \emph{tail(A,C)} forms a \emph{minimal unsatisfiable subprogram} (MUSP) of $h$ as each of the literals alone is satisfiable.

Existing ILP approaches do not identify MUSPs.
Some approaches reason about the failure of an entire program but cannot explain precisely \emph{why} a program fails \cite{hexmil,popper,hopper}.
Some approaches identify erroneous rules \cite{mis,prosynth} or 
literals \cite{mugg:metagold,popperx} but do not identify MUSPs.

To overcome this limitation, we introduce an approach that identifies MUSPs.
To explore our idea, we use the \emph{learning from failures} (LFF) setting \cite{popper}.
LFF frames the ILP problem as a constraint satisfaction problem (CSP) where each solution to the CSP represents a program.
The goal of a LFF learner is to accumulate constraints to restrict the hypothesis space.
We build on the LFF system \popper{} \cite{popper}, which can learn optimal and recursive programs from infinite domains. 
We enable \popper{} to identify MUSPs and build constraints from them to guide subsequent search.
We call our system \name{}.

\subsubsection*{Novelty and Contributions}
The main novelty of this paper is \emph{identifying minimal unsatisfiable subprograms} and building constraints from them.
The impact, which we demonstrate on multiple domains, is vastly improved learning performance.
Moreover, as the idea connects many areas of AI, including machine learning, logic programming, and constraint satisfaction, we hope the idea interests a broad audience.
Overall, we make the following contributions:

\begin{itemize}
\item We describe the \emph{minimal unsatisfiable subprogram} problem, the problem of finding MUSPs.
We show that pruning with MUSPs is optimally sound (Propositions \ref{prop_spec_sound} and \ref{prop_redund_sound}) and effective (Propositions \ref{prop_spec} and \ref{prop_redund}).

\item We implement our approach in \name{}, an ILP system which identifies the MUSPs of definite programs, including recursive programs.
We prove that \name{} always learns an optimal solution if one exists (Theorem \ref{thm:optcorrect}).

\item We experimentally show on multiple domains, including program synthesis and game playing, that our approach can substantially reduce learning times by 99\%.
\end{itemize}
\section{Related Work}
\label{sec:related}

\textbf{Rule mining.}
ILP is a form of rule mining.
Comparing our approach with rule mining approaches is difficult.
For instance, AMIE+ \cite{DBLP:journals/vldb/GalarragaTHS15}, a rule mining system, uses an open-world assumption, can only use unary and binary relations, and requires facts as input.
By contrast, \name{} uses the closed-world assumption, can use any arity relations, and definite programs as BK.

\textbf{ILP.}
Many ILP systems \cite{progol,tilde,aleph,quickfoil,probfoil} struggle to learn recursive programs and perform predicate invention \cite{ilp30}.
By contrast, \name{} can learn recursive programs and perform predicate invention.
In addition, unlike many ILP systems \cite{dilp,hexmil,DBLP:conf/sigsoft/SiLZAKN18,meta_abduce,DBLP:conf/icml/GlanoisJFWZ0LH22}, we do not require metarules.

\textbf{Rule selection.}
Many systems formulate the ILP problem as a rule selection problem \cite{aspal,ilasp3,hexmil,prosynth,shitruleselection}.
These approaches precompute every possible rule in the hypothesis space and then search for a subset that entails the examples.
Because they precompute all possible rules, they cannot learn rules with many literals.
By contrast, we do not precompute every possible rule.
Moreover, we can learn Datalog and definite programs.

\textbf{Theory repair and revision.}
Theory repair approaches \cite{bundy2016reformation} revise a hypothesis by applying generalisation and specialisation operators to revision points.
Theory revision approaches \cite{DBLP:journals/ml/Wrobel94,ruth,DBLP:journals/ml/PaesZC17} can identify literals as revision points, but often with limitations, such as requiring user-interaction \cite{mis} or restrictions on the training examples \cite{forte}.
By contrast, \name{} is guaranteed to find an optimal solution if one exists without such restrictions.


\textbf{Explanations.}
\popper{} \cite{popper}  builds constraints from whole programs to prune the hypothesis space.
By contrast, we build constraints from MUSPs, which leads to better pruning (Section \ref{sec:benefits}).
\textsc{ilasp3} \cite{ilasp3} and \textsc{ProSynth} \cite{prosynth} generate constraints from rules that do not entail an example.
These two approaches precompute all possible rules and only identify erroneous rules. 
By contrast, we do not precompute all possible rules and identify minimal subsets of unsatisfiable literals.
\metagol{} \cite{mugg:metagold} builds a program literal-by-literal and tests it on the examples.
If a program does not entail any positive examples, \metagol{} backtracks to consider alternative literals, i.e. it identifies unsatisfiable literals.
\metagol{} repeatedly reconsiders unsatisfiable programs, whereas \name{} learns constraints so it never does.
\textsc{Hempel} \cite{popperx} analyses the execution paths of definite programs to identify unsatisfiable literals.
\textsc{Hempel} analyses rules using SLD-resolution from left to right and, therefore, can only identify definite subprograms and is not guaranteed to find MUSPs.
For instance, given the unsatisfiable rule 
\emph{f(A,B) $\leftarrow$ element(A,B),odd(B),even(B)},
\textsc{Hempel} cannot find a smaller unsatisfiable subprogram.
By contrast, \name{} identifies that the literals \emph{odd(B)} and \emph{even(B)} form a MUSP.

        


\textbf{Constraint discovery.}
Clausal discovery approaches \cite{claudien} discover constraints to include in a hypothesis to eliminate models.
By contrast, we discover constraints to prune the hypothesis space.
\disco{} \cite{disco} discovers constraints from the BK as a preprocessing step.
It uses a predefined set of constraints, such as \emph{asymmetry} and \emph{antitransitivity}.
We differ because we (i) find MUSPs, (ii) discover constraints during the learning, not as a preprocessing step, (iii) do not use a predefined set of constraints, and (iv) jointly consider the BK and the examples.
As our experiments show, these differences greatly impact learning performance.


\section{Problem Setting}
\label{sec:setting}

We describe our problem setting. 
We assume familiarity with logic programming \cite{lloyd:book} but have included a summary in the appendix.

\subsection{Learning From Failures}
We use the LFF setting.
We learn definite programs with the least Herbrand model semantics.
A \emph{hypothesis} is a set of definite clauses.
We use the term \emph{program} synonymously with \emph{hypothesis}.
A \emph{hypothesis space} $\mathcal{H}$ is a set of hypotheses.
LFF uses \emph{hypothesis constraints} to restrict the hypothesis space.
Let $\mathcal{L}$ be a language that defines hypotheses.
A \emph{hypothesis constraint} is a constraint (a headless rule) expressed in $\mathcal{L}$.
Let $C$ be a set of hypothesis constraints written in a language $\mathcal{L}$.
A hypothesis is \emph{consistent} with $C$ if when written in $\mathcal{L}$ it does not violate any constraint in $C$.
We denote as $\mathcal{H}_{C}$ the set of hypotheses from $\mathcal{H}$ which are consistent with $C$.

We define the LFF input:

\begin{definition}[\textbf{LFF input}]
\label{def:probin}
A \emph{LFF} input is a tuple $(E^+, E^-, B, \mathcal{H}, C)$ where $E^+$ and $E^-$ are sets of ground atoms denoting positive and negative examples respectively; $B$ is a definite program denoting background knowledge;
$\mathcal{H}$ is a hypothesis space, and $C$ is a set of hypothesis constraints.
\end{definition}

\noindent
We define a LFF solution:

\begin{definition}[\textbf{LFF solution}]
\label{def:solution}
Given a LFF input $(E^+, E^-, B, \mathcal{H}, C)$, a hypothesis $h \in \mathcal{H}_{C}$ is a \emph{solution} when $h$ is \emph{complete} ($\forall e \in E^+, \; B \cup h \models e$) and \emph{consistent} ($\forall e \in E^-, \; B \cup h \not\models e$).
\end{definition}

\noindent
We define an \emph{optimal} solution, where \emph{size(h)} denote the number of literals in the hypothesis $h$:

\begin{definition}
[\textbf{Optimal solution}]
\label{def:opthyp}
Given a LFF input $(E^+, E^-, B, \mathcal{H}, C)$, a hypothesis $h \in \mathcal{H}_{C}$ is \emph{optimal} when (i) $h$ is a solution, and (ii) $\forall h' \in \mathcal{H}_{C}$, where $h'$ is a solution, $size(h) \leq size(h')$.
\end{definition}

\noindent
If a hypothesis is not a solution then it is a \emph{failure}.
A hypothesis is \emph{incomplete} when $\exists e \in E^+, \; h \cup B \not \models e$; 
\emph{inconsistent} when $\exists e \in E^-, \; h \cup B \models e$;
\emph{partially complete} when $\exists e \in E^+, \; h \cup B \models e$; 
and \emph{totally incomplete} when $\forall e \in E^+, \; h \cup B \not \models e$.



\paragraph{Constraints.}
A LFF learner builds constraints from failures to prune the hypothesis space.
Cropper and Morel \shortcite{popper} introduce constraints based on subsumption \cite{plotkin:thesis}.
A clause $c_1$ \emph{subsumes} a clause $c_2$ ($c_1 \preceq c_2$) if and only if there exists a substitution $\theta$ such that $c_1\theta \subseteq c_2$.
A clausal theory $t_1$ subsumes a clausal theory $t_2$ ($t_1 \preceq t_2$) if and only if $\forall c_2 \in t_2, \exists c_1 \in t_1$ such that $c_1$ subsumes $c_2$.
A clausal theory $t_1$ is a \emph{specialisation} of a clausal theory $t_2$ if and only if $t_2 \preceq t_1$.
A clausal theory $t_1$ is a \emph{generalisation} of a clausal theory $t_2$ if and only if $t_1 \preceq t_2$.
There are three standard constraints for a failure hypothesis $h$. 
A \emph{specialisation} constraint prunes specialisations of $h$.
A \emph{generalisation} constraint prunes generalisations of $h$.
Morel and Cropper \shortcite{poppi} introduce \emph{redundancy} constraints, which are difficult to tersely describe.
Although our algorithm supports predicate invention, for simplicity, in the following high-level explanation, we ignore predicate invention and focus on redundancy constraints for recursive programs.
If a hypothesis $h$ is totally incomplete then $h$ (or a specialisation of it) cannot appear in an optimal solution $h_o$ unless $h_o$ has a recursive rule that does not specialise a rule in $h$.
For instance, if a totally incomplete hypothesis has a single rule $r$ then $r$ can only be in an optimal solution $h_o$ if $h_o$ has a recursive rule that does not specialise $r$. Otherwise, $r$ is redundant in $h_o$.
If every recursive rule in a hypothesis $p$ specialises a rule in a hypothesis $q$ then we say that $p$ is \emph{redundant} with respect to $q$.
The appendix includes examples of redundant programs.
\subsection{Minimal Unsatisfiable Subprograms}
We define the MUSP problem.
We define a \emph{subprogram}:


\begin{definition}[\textbf{Subprogram}]
\label{def:core}
    Let $h$ be a hypothesis and $g$ be a set of Horn clauses.
    Then $g$ is a \emph{subprogram} of $h$ if and only if the following conditions all hold:
    (i) $size(g) < size(h)$,
    (ii) $g \preceq h$, 
    and
    (iii) $\forall c_1 \in g, \exists c_2 \in h$ such that $c_1 \preceq c_2$.
\end{definition}

\noindent
\begin{example}
Consider the program:
\[
    \begin{array}{l}
    h = \left\{
    \begin{array}{l}
        \emph{f(A,B) $\leftarrow$ tail(A,C), empty(C), head(A,B)}\\    
        \emph{f(A,B) $\leftarrow$ tail(A,C), f(C,B)}
    \end{array}
    \right\}
    \end{array}
\]
Then $h$ has the subprogram:
\[
    \begin{array}{l}
    \left\{
    \begin{array}{l}
        \emph{f(A,B) $\leftarrow$ tail(A,C), head(A,B)}\\    
        \emph{f(A,B) $\leftarrow$ tail(A,C), f(C,B)}
    \end{array}
    \right\}
    \end{array}
\]
\noindent
The following program is not a subprogram of $h$ as it has more literals than $h$ and thus violates condition (i):
\[
    \begin{array}{l}
    \left\{
    \begin{array}{l}
        \emph{f(A,B) $\leftarrow$ tail(A,C), tail(C,D), head(A,B)}\\    
        \emph{f(A,B) $\leftarrow$ tail(A,C), empty(C), head(A,B)}\\    
        \emph{f(A,B) $\leftarrow$ tail(A,C), f(C,B)}
    \end{array}
    \right\}
    \end{array}
\]
\noindent
The following program is not a subprogram of $h$ as it does not subsume $h$ and thus violates condition (ii):
\[
    \begin{array}{l}
    \left\{
    \begin{array}{l}
        \emph{f(A,B) $\leftarrow$ tail(A,C), empty(C), head(A,B)}
    \end{array}
    \right\}
    \end{array}
\]
The following program is not a subprogram of $h$ as its second rule does not subsume a rule in $h$ and thus it violates condition (iii):
\[
    \begin{array}{l}
    \left\{
    \begin{array}{l}
        \emph{f(A,B) $\leftarrow$ tail(A,C)}\\    
        \emph{f(A,B) $\leftarrow$ empty(A), tail(A,C)}
    \end{array}
    \right\}
    \end{array}
\]
\end{example}

\noindent
We define an \emph{unsatisfiable subprogram}:

\begin{definition}[\textbf{Unsatisfiable subprogram}]
\label{def:uc}
Let $(E^+, E^-, B, \mathcal{H}, C)$ be a LFF input, 
$h \in \mathcal{H}_{C}$ be a hypothesis, $g$ be a subprogram of $h$, and $g$ be totally incomplete with respect to $E^+$ and $B$.
Then $g$ is an \emph{unsatisfiable subprogram} of $h$.
\end{definition}

\begin{example}
\label{ex:uc}
\noindent
Consider the positive examples $E^+=\{f([],0), f([i,j,c,a,i],5)\}$ and the hypothesis:
\[
    \begin{array}{l}
    h = \left\{
    \begin{array}{l}
        \emph{f(A,B) $\leftarrow$ empty(A), head(A,B), tail(A,C), head(C,B)}
    \end{array}
    \right\}
    \end{array}
\]
Assuming standard definitions for the relations, two unsatisfiable subprograms of $h$ are:
\[
    \begin{array}{l}
    \left\{
    \begin{array}{l}
        \emph{f(A,B) $\leftarrow$ empty(A), head(A,B)}
    \end{array}
    \right\}
    \\
    \left\{
    \begin{array}{l}
        \emph{$\leftarrow$ empty(A), head(A,B), tail(A,C)}
    \end{array}
    \right\}
    \end{array}
\]
\end{example}

\noindent
Note that a subprogram can contain headless rules (goal clauses) as in the second subprogram above.

We define a \emph{minimal unsatisfiable subprogram}:
\begin{definition}[\textbf{Minimal unsatisfiable subprogram}]
\label{def:muc}
Let $S$ be the set of all unsatisfiable subprograms of the hypothesis $h$.
Then $g \in S$ is a \emph{minimal unsatisfiable subprogram} (MUSP) of $h$ if and only if
$\forall g' \in S, size(g) \leq size(g')$.
\end{definition}

\begin{example}
\noindent
Reusing $h$ and $E^+$ from Example \ref{ex:uc}, $h$ has the MUSPs: 
\[
    \begin{array}{l}
    \left\{
    \begin{array}{l}
        \emph{$\leftarrow$ empty(A), head(A,B)}
    \end{array}
    \right\}
    \\
    \left\{
    \begin{array}{l}
        \emph{$\leftarrow$ empty(A), tail(A,C)}
    \end{array}
    \right\}
    \\
    \left\{
    \begin{array}{l}
        \emph{f(A,B) $\leftarrow$ head(A,B)}
    \end{array}
    \right\}
    \end{array}
\]
\end{example}



\noindent
A MUSP differs to a minimal unsatisfiable core \cite{unsat_core} and a minimal unsatisfiable subset \cite{DBLP:journals/ai/AlvianoDFPR23}, which are both usually defined in terms of subset minimality.
By contrast, a MUSP is defined in terms of cardinality minimality.

We define the \emph{MUSP problem}:
\begin{definition}[\textbf{MUSP problem}]
\label{def:muc-prob}
The \emph{MUSP problem} is to find all MUSPs of a hypothesis.
\end{definition}

\noindent
Section \ref{sec:algo} describes \name{}, which solves the MUSP problem.




\subsection{MUSP Constraints}
\label{sec:cons}
We want to identify MUSPs to build constraints from them to prune the hypothesis space of an ILP learner.
We describe the constraints that we can build from a MUSP.
Due to space limitations, all the proofs are in the appendix.

Since a MUSP is totally incomplete then its specialisations cannot be solutions:

\begin{proposition}[\textbf{MUSP specialisations}]
Let $h$ be a program, $m$ be a MUSP of $h$, and $h'$ be a specialisation of $m$.
Then $h'$ is not a solution.
\label{prop_spec_sound}
\end{proposition}

\noindent
We show an example of this constraint:
\begin{example}
\label{ex:specialisation}
Consider the positive example \emph{$E^+$ = \{f([x],y)\}} and the hypothesis:
\[
    \begin{array}{ll}
    h=& \left\{
    \begin{array}{l}
        \emph{f(A,B) $\leftarrow$ tail(A,C), tail(C,A), head(C,B)}
    \end{array}
    \right\}
    \end{array}
\]
A MUSP of $h$ is:
\[
    \begin{array}{ll}
    m = 
    & \left\{
    \begin{array}{l}
      \emph{$\leftarrow$ tail(A,C), tail(C,A)}
    \end{array}
    \right\}
    \end{array}
\]
We can prune the specialisations of $m$, such as:
\[
    \begin{array}{l}
    \left\{
    \begin{array}{l}
      \emph{f(A,B) $\leftarrow$ tail(A,C), tail(C,A)}
    \end{array}
    \right\}\\
    \left\{
    \begin{array}{l}
      \emph{$\leftarrow$ tail(A,C), tail(C,A), head(A,D)}
    \end{array}
    \right\}
    \end{array}
\]
\end{example}

\noindent
Since a MUSP $m$ is totally incomplete then its specialisations cannot be in an optimal solution without a recursive rule that does not specialise $m$:

\begin{proposition}[\textbf{MUSP redundancy}]
Let $h$ and $h'$ be hypotheses, $m$ be a MUSP of $h$, and $h'$ be redundant with respect to $m$.
Then $h'$ is not an optimal solution.
\label{prop_redund_sound}
\end{proposition}

\noindent
We show an example of this constraint:

\begin{example}
Consider the positive example \emph{$E^+$ = \{f([x],y)\}} and the hypothesis:
\[
    \begin{array}{ll}
    h=& \left\{
    \begin{array}{l}
        \emph{f(A,B) $\leftarrow$ tail(A,C), tail(C,D), head(D,B)}
    \end{array}
    \right\}
    \end{array}
\]
A MUSP of $h$ is:
\[
    \begin{array}{ll}
    m = 
    & \left\{
    \begin{array}{l}
      \emph{f(A,B) $\leftarrow$ tail(A,C), tail(C,D)}
    \end{array}
    \right\}
    \end{array}
\]
We can prune the following hypothesis because it is redundant with respect to $m$:
\[
    \begin{array}{l}
    \left\{
    \begin{array}{l}
      \emph{f(A,B) $\leftarrow$ head(A,B)}\\
      \emph{f(A,B) $\leftarrow$ tail(A,C), tail(C,D), f(D,B)}
    \end{array}
    \right\}
    \end{array}
\]
\end{example}
\noindent
In Section \ref{sec:algo}, we introduce \name{} which uses these optimally sound constraints to prune the hypothesis space.

\subsection{MUSP Benefits}
\label{sec:benefits}
We show that using MUSPs to build constraints that prune specialisations and redundant hypotheses leads to more pruning. 
The proofs are in the appendix.
\begin{proposition}[\textbf{Specialisation pruning}] 
\label{prop_spec} 
Let $h$ be a hypothesis, $m$ be a MUSP of $h$, $S_h$ be the specialisations of $h$, and $S_m$ be the specialisations of $m$.
Then $S_h \subset S_m$.
\end{proposition}

\begin{proposition}[\textbf{Redundancy pruning}] 
Let $h$ be a hypothesis,
$m$ be a MUSP of $h$, 
$R_h$ be the hypotheses redundant with respect to $h$,
and $R_m$ be the hypotheses redundant with respect to $m$.
Then $R_h \subset R_m$.
\label{prop_redund}
\end{proposition}
\section{Algorithm}
\label{sec:algo}

We now describe our \name{} approach.
To delineate the novelty, we first describe \popper{}.


\paragraph{\popper{}.}

\noindent
\popper{} takes as input background knowledge (\emph{bk}), positive (\emph{pos}) and negative (\emph{neg}) training examples, and a maximum hypothesis size (\emph{max\_size}).
\popper{} uses a generate, test, and constrain loop to find an optimal solution.
\popper{} starts with a CSP program $\mathcal{P}$ (hidden in the generate function).
The models of $\mathcal{P}$ correspond to hypotheses (definite programs).
In the generate stage, \popper{} 
searches for a model of $\mathcal{P}$.
If there is no model, \popper{} increments the hypothesis size and loops again.
If there is a model, \popper{} converts it to a hypothesis $h$.
In the test stage, \popper{} uses Prolog to test $h$ on the examples.
If $h$ is a solution, \popper{} returns it.
If $h$ is a failure then, in the constrain stage, \popper{} builds hypothesis constraints (represented as CSP constraints) from $h$.
\popper{} adds these constraints to $\mathcal{P}$ to prune models and thus prune the hypothesis space.
For instance, if $h$ is incomplete, \popper{} builds a specialisation constraint to prune its specialisations.
If $h$ is inconsistent, \popper{} builds a generalisation constraint to prune its generalisations. 
To reiterate, \popper{} builds constraints from whole programs.
\popper{} repeats this loop until it finds an optimal solution or exhausts the models of $\mathcal{P}$.


      
\subsection*{\name{}}
\name{} (Algorithm \ref{alg:explainer}) is similar to \popper{} except for one major difference.
If a program $h$ is totally incomplete then \name{}  calls Algorithm \ref{alg:explain_incomplete} (line 13) to identify the MUSPs of $h$ and build constraints from them.
To reiterate, the ability to identify MUSPs and build constraints from them is the main novelty and contribution of this paper.

\begin{algorithm}[ht!]
\small
{
\begin{myalgorithm}[]
def MUSPer(bk, pos, neg, max_size):
  cons = {}
  size = 1
  while size $\leq$ max_size:
    h = generate(cons, size)
    if h == UNSAT:
      size += 1
      continue
    outcome = test(pos, neg, bk, h)
    if outcome == (COMPLETE, CONSISTENT)
      return h
    if outcome == (TOTALLY_INCOMPLETE, _)
      cons += unsat_constraints(h, pos, bk)    
    cons += constrain(h, outcome)
  return {}
\end{myalgorithm}
\caption{
\name{}
}
\label{alg:explainer}
}
\end{algorithm}

\begin{algorithm}[ht!]
\small
{
\begin{myalgorithm}[]
def unsat_constraints(h, pos, bk):
  MUSPs = find_MUSPs(h, pos, bk)
  cons = {}
  for MUSP in MUSPs:
    cons += build_spec_con(MUSP)
    cons += build_redund_con(MUSP)
  return cons

def find_MUSPs(h, pos, bk):
  MUSPs = {}
  has_unsat_subprog = False
  for subprog in subprogs(h):
    if unsat(subprog, pos, bk): 
       has_unsat_subprog = True
       MUSPs += find_MUSPs(subprog, pos, bk):
  if has_unsat_subprog == False:
    MUSPs += {h}
  return MUSPs
\end{myalgorithm}
\caption{
Find MUSPs
}
\label{alg:explain_incomplete}
}
\end{algorithm}


\subsubsection{MUSPs}
Algorithm \ref{alg:explain_incomplete} builds constraints from MUSPs.
To find MUSPs (line 2), we use what Dershowitz et al. \shortcite{DBLP:conf/sat/DershowitzHN06} call a \emph{folk algorithm}.
We search for subprograms of the program $h$ by removing a literal from it and checking that the subprogram obeys the conditions in Definition \ref{def:core} (line 12).
We also enforce other syntactic restrictions, such as that the literals in a rule are connected\footnote{The literals in a rule are connected if they cannot be partitioned into two sets such that the variables in the literals of one set are disjoint from the variables in the literals of the other set.}.
We check each subprogram to see if it is unsatisfiable (line 13).
If a subprogram is unsatisfiable, we recursively find the MUSPs of the subprogram itself and add the result to a set of minimal unsatisfiable subprograms (line 15).
Otherwise, we add $h$ to the set of minimal unsatisfiable subprograms (line 17).
\subsubsection{Satisfiability}
In our implementation of Algorithm \ref{alg:explain_incomplete} we use Prolog to check the satisfiability of a subprogram (line 13).
There are two cases.
If all the rules in a subprogram $s$ have a head literal then we say that $s$ is satisfiable if it entails at least one positive example.
If a subprogram $s$ contains a rule $r$ that does not have a head literal then we add a \emph{is\_sat} literal to the head of $r$. We say that $s$ is satisfiable if \emph{is\_sat} is true.
For instance, given the subprogram:
\[
    \begin{array}{ll}
    & \left\{
    \begin{array}{l}
      \emph{$\leftarrow$ tail(A,C), tail(C,A)}
    \end{array}
    \right\}
    \end{array}
\]
We test the program:
\[
    \begin{array}{ll}
    & \left\{
    \begin{array}{l}
      \emph{is\_sat $\leftarrow$ tail(A,C), tail(C,A)}
    \end{array}
    \right\}
    \end{array}
\]

\subsubsection{Constraints}
Since a MUSP is totally incomplete then its specialisations cannot be a solution (Proposition \ref{prop_spec_sound}).
Likewise, since a MUSP is totally incomplete then we can prune programs redundant with respect to it (Proposition \ref{prop_redund_sound}). 
Therefore, Algorithm \ref{alg:explain_incomplete} builds a specialisation and a redundancy constraint for each MUSP (lines 5 and 6).

\subsubsection{Correctness}

We show that \name{} is correct:

\begin{theorem}[\textbf{\name{} correctness}]
\label{thm:optcorrect}
\name{} returns an optimal solution if one exists.
\end{theorem}
\begin{proof}
Cropper and Morel \shortcite{popper} show that given optimally sound constraints \popper{} returns an optimal solution if one exists (Theorem 1). \name{} builds on \popper{} by additionally pruning specialisations and redundancies of MUSPs, which are optimally sound by Propositions \ref{prop_spec_sound} and \ref{prop_redund_sound} respectively. 
Therefore, \name{} never prunes optimal solutions so returns an optimal solution if one exists.
\end{proof}


\section{Experiments}
\label{sec:exp}
We claim that identifying MUSPs can improve the learning performance of an ILP system.
Propositions \ref{prop_spec} and \ref{prop_redund} support this claim and show that identifying MUSPs improves pruning.
However, it is unclear whether in practice the overhead of finding MUSPs and building the constraints outweigh the pruning benefits.
Therefore, our experiments aim to answer the question:
\begin{description}
\item[Q1] Can identifying MUSPs reduce learning times?
\end{description}

\noindent
To answer \textbf{Q1}, we compare the performance of \name{} against \popper{}.
As \name{} builds on \popper{}, the only difference between the systems is the ability to identify MUSPs, so this comparison directly measures the impact of our idea.

To understand how \name{} compares to other approaches, our experiments try to answer the question:
\begin{description}
\item[Q2] How does \name{} compare to other approaches?
\end{description}
\noindent
To answer \textbf{Q2}, we compare \name{} against \ale{} \cite{aleph}, \metagol{} \cite{mugg:metagold}, \popper{}, and \disco{} \cite{disco}\footnote{
We tried other systems but few are usable on our domains.
For instance, rule selection approaches (Section \ref{sec:related}) precompute all possible rules which is infeasible on our domains and would, for example, require precomputing at least $10^{15}$ rules for the \emph{coins-goal} task.
}.





\paragraph{Reproducibility.} 
All the experimental code and data are included as supplementary material and will be made publicly available if the paper is accepted for publication.

\subsection{Domains}

We briefly describe our domains.
The appendix contains more details, such as example solutions for each task and statistics about the problem sizes.

\textbf{Trains.}
The goal is to find a hypothesis that distinguishes east and west trains \cite{michalski:trains}.

\textbf{Zendo.}
Zendo is a multiplayer game where players must discover a secret rule by building structures.
Zendo has attracted interest in cognitive science \cite{zendo}.

\textbf{IMDB.}
We use a frequently used real-world dataset which contains relations between movies, actors, and directors \cite{imdb,alps}.

\textbf{IGGP.}
The goal of \emph{inductive general game playing} \cite{iggp} (IGGP) is to induce rules to explain game traces from the general game playing competition \cite{ggp}.

\textbf{Chess.} 
The task is to learn chess patterns in the king-rook-king (\emph{krk}) endgame \cite{celine:bottom}.


\textbf{Program synthesis.}
We use a program synthesis dataset \cite{popper} augmented with 5 new tasks. 


\textbf{SQL.} This dataset \cite{DBLP:conf/sigsoft/SiLZAKN18} contains 15 real-world tasks that involve synthesising SQL queries.
As there is no testing data, we report training accuracy.


\subsection{Experimental Setup}
We enforce a timeout of 30 minutes per task.
The appendix includes all the experimental details and example solutions.
We measure predictive accuracy, learning time (including the time to discover MUSPs), and the number of programs considered. 
We also report the time taken to discover MUSPs, i.e. the overhead of our approach. 
We measure the mean over 10 trials.
The error bars in all tables denote standard error.
\subsection{Experimental Results}

\subsubsection{\textbf{Q1.} Can Identifying MUSs Reduce Learning Times?}
Table \ref{tab:qtimes} shows the learning times aggregated per domain. 
Table \ref{tab:qtimesall} shows the learning times for tasks where the times of \name{} and \popper{} differ.
These results show that \name{} never needs more time than \popper{}.
A paired t-test confirms the significance of the difference ($p < 0.01$). 
\name{} can drastically reduce learning time. 
For instance, on the \emph{sql-06} task, \popper{} times out after 1800s, whilst \name{} takes only 14s, a 99\% reduction. 
Likewise, on the \emph{coins-g} task, \popper{} times out after 1800s whilst \name{} only takes 6s, a 99\% reduction.

Table \ref{tab:qaccs} shows the predictive accuracies aggregated per domain.
\name{} has equal or higher predictive accuracy than \popper{} on all but one domain, where the difference is negligible. 
These results show that \name{} can drastically reduce learning times whilst maintaining high levels of predictive accuracy.


\begin{table}[ht!]
\centering
\begin{tabular}{@{}l|ccc@{}}
    \textbf{Domain} & \textbf{\popper{}} & \textbf{\name{}} & \textbf{Change}\\
    \midrule
\emph{trains} & 13 $\pm$ 1 & 13 $\pm$ 1 & 0\% \\
\emph{zendo} & 36 $\pm$ 4 & 36 $\pm$ 4 & 0\% \\
\emph{imdb} & 193 $\pm$ 51 & 148 $\pm$ 39 & \textbf{--23\%} \\
\emph{iggp} & 617 $\pm$ 66 & 207 $\pm$ 34 & \textbf{--66\%} \\
\emph{krk} & 41 $\pm$ 6 & 38 $\pm$ 5 & \textbf{--7\%} \\
\emph{synthesis} & 343 $\pm$ 39 & 199 $\pm$ 33 & \textbf{--41\%} \\
\emph{sql} & 594 $\pm$ 63 & 13 $\pm$ 1 & \textbf{--97\%} \\
\end{tabular}
\caption{
Learning times (seconds).
}
\label{tab:qtimes}
\end{table}


{
\setlength{\tabcolsep}{0.2em} 
\begin{table}[ht!]
\centering
\footnotesize
\begin{tabular}{@{}l|ccccc@{}}
    \textbf{Domain} &  
    \textbf{\ale{}} &
    \textbf{\metagol{}} &
    \textbf{\disco{}} & 
    \textbf{\popper{}} &
    \textbf{\name{}}\\
    \midrule
\emph{trains} & 100 $\pm$ 0 & 45 $\pm$ 4 & 99 $\pm$ 0 & 99 $\pm$ 0 & 99 $\pm$ 0 \\
\emph{zendo} & 93 $\pm$ 1 & 55 $\pm$ 2 & 97 $\pm$ 1 & 96 $\pm$ 1 & 97 $\pm$ 1 \\
\emph{imdb} & 67 $\pm$ 4 & 39 $\pm$ 3 & 100 $\pm$ 0 & 100 $\pm$ 0 & 100 $\pm$ 0 \\
\emph{iggp} & 88 $\pm$ 3 & 34 $\pm$ 3 & 99 $\pm$ 0 & 100 $\pm$ 0 & 99 $\pm$ 0 \\
\emph{krk} & 95 $\pm$ 1 & 50 $\pm$ 0 & 70 $\pm$ 4 & 70 $\pm$ 4 & 73 $\pm$ 4 \\
\emph{synthesis} & 54 $\pm$ 1 & 57 $\pm$ 1 & 97 $\pm$ 1 & 97 $\pm$ 1 & 98 $\pm$ 1 \\
\emph{sql} & 100 $\pm$ 0 & 13 $\pm$ 3 & 95 $\pm$ 2 & 88 $\pm$ 3 & 100 $\pm$ 0 \\
\end{tabular}
\caption{
Predictive accuracies (\%).
}
\label{tab:qaccs}
\end{table}
}

\begin{table}[ht!]
\centering
\begin{tabular}{@{}l|ccc@{}}
    \textbf{Task} & \textbf{\popper{}} & \textbf{\name{}} & \textbf{Change}\\
\midrule
\emph{zendo3} & 48 $\pm$ 6 & \textbf{46 $\pm$ 4}  & \textbf{-4\%} \\
\midrule
\emph{imdb3} & 571 $\pm$ 37 & \textbf{435 $\pm$ 34}  & \textbf{-23\%} \\
\midrule
\emph{md} & 270 $\pm$ 25 & \textbf{53 $\pm$ 2}  & \textbf{-80\%} \\
\emph{buttons} & 743 $\pm$ 144 & \textbf{50 $\pm$ 4}  & \textbf{-93\%} \\
\emph{rps} & 185 $\pm$ 2 & \textbf{162 $\pm$ 2}  & \textbf{-12\%} \\
\emph{coins} & 1028 $\pm$ 71 & \textbf{952 $\pm$ 63}  & \textbf{-7\%} \\
\emph{buttons-g} & 21 $\pm$ 0.5 & \textbf{12 $\pm$ 0}  & \textbf{-42\%} \\
\emph{coins-g} & \emph{timeout} & \textbf{6 $\pm$ 0}  & \textbf{-99\%} \\
\emph{attrition} & 740 $\pm$ 11 & \textbf{290 $\pm$ 24}  & \textbf{-60\%} \\
\emph{centipede} & 147 $\pm$ 2 & \textbf{129 $\pm$ 11}  & \textbf{-12\%} \\
\midrule
\emph{krk} & 41 $\pm$ 6 & \textbf{38 $\pm$ 5}  & \textbf{-7\%} \\
\midrule
\emph{dropk} & 10 $\pm$ 1 & \textbf{7 $\pm$ 1}  & \textbf{-30\%} \\
\emph{droplast} & 27 $\pm$ 2 & \textbf{19 $\pm$ 1}  & \textbf{-29\%} \\
\emph{evens} & 58 $\pm$ 7 & \textbf{38 $\pm$ 3}  & \textbf{-34\%} \\
\emph{finddup} & 133 $\pm$ 12 & \textbf{74 $\pm$ 11}  & \textbf{-44\%} \\
\emph{last} & 22 $\pm$ 2 & \textbf{11 $\pm$ 1}  & \textbf{-50\%} \\
\emph{contains2} & 306 $\pm$ 11 & \textbf{268 $\pm$ 12}  & \textbf{-12\%} \\
\emph{length} & 57 $\pm$ 6 & \textbf{24 $\pm$ 5}  & \textbf{-57\%} \\
\emph{reverse} & 1053 $\pm$ 149 & \textbf{203 $\pm$ 19}  & \textbf{-80\%} \\
\emph{sorted} & 253 $\pm$ 25 & \textbf{216 $\pm$ 26}  & \textbf{-14\%} \\
\emph{sumlist} & 276 $\pm$ 49 & \textbf{25 $\pm$ 1}  & \textbf{-90\%} \\
\emph{next} & 59 $\pm$ 10 & \textbf{44 $\pm$ 7}  & \textbf{-25\%} \\
\emph{rotateN} & 437 $\pm$ 43 & \textbf{292 $\pm$ 49}  & \textbf{-33\%} \\
\emph{inttobin} & 83 $\pm$ 12 & \textbf{65 $\pm$ 8}  & \textbf{-21\%} \\
\emph{chartointallodd} & 638 $\pm$ 149 & \textbf{104 $\pm$ 5}  & \textbf{-83\%} \\
\emph{twosucc} & \emph{timeout} & \textbf{1764 $\pm$ 23}  & \textbf{-2\%} \\
\midrule
\emph{sql-01} & 14 $\pm$ 6 & \textbf{6 $\pm$ 0}  & \textbf{-57\%} \\
\emph{sql-02} & 936 $\pm$ 136 & \textbf{29 $\pm$ 1}  & \textbf{-96\%} \\
\emph{sql-03} & 1126 $\pm$ 138 & \textbf{24 $\pm$ 2}  & \textbf{-97\%} \\
\emph{sql-04} & \emph{timeout} & \textbf{71 $\pm$ 4}  & \textbf{-96\%} \\
\emph{sql-05} & 8 $\pm$ 0 & \textbf{5 $\pm$ 0}  & \textbf{-37\%} \\
\emph{sql-06} & \emph{timeout} & \textbf{14 $\pm$ 1}  & \textbf{-99\%} \\
\emph{sql-07} & 9 $\pm$ 0 & \textbf{6 $\pm$ 0}  & \textbf{-33\%} \\
\emph{sql-08} & 1527 $\pm$ 142 & \textbf{8 $\pm$ 0}  & \textbf{-99\%} \\
\emph{sql-09} & 5 $\pm$ 0 & \textbf{4 $\pm$ 0}  & \textbf{-20\%} \\
\emph{sql-10} & 11 $\pm$ 1 & \textbf{6 $\pm$ 0}  & \textbf{-45\%} \\
\emph{sql-11} & 1615 $\pm$ 125 & \textbf{8 $\pm$ 0}  & \textbf{-99\%} \\
\emph{sql-12} & 41 $\pm$ 11 & \textbf{5 $\pm$ 0}  & \textbf{-87\%} \\
\end{tabular}
\caption{
Learning times (seconds). 
We only show tasks where the two approaches differ.
The full table is in the appendix.
}
\label{tab:qtimesall}
\end{table}



Table \ref{tab:num_progs} shows the aggregated number of programs considered by each system. 
The full table for each task is in the appendix.
\name{} can consider considerably fewer programs than \popper{}.
For instance, on the \emph{sql-02} task, \popper{} considers 25,533 programs whereas \name{} only considers 52, a 99\% reduction.
This result shows that constraints built from MUSPs can effectively prune the hypothesis space. 

Table \ref{tab:overhead} shows the proportion of the total learning time spent identifying MUSPs. 
The overhead is usually less than a few seconds and represents at most 23\% of the learning time. 

To illustrate why our approach works, consider the \emph{last} task, where \name{} can generate the hypothesis:
\[
    \begin{array}{l}
    \left\{
    \begin{array}{l}
        \emph{f(A,B) $\leftarrow$ head(A,B), zero(C), decrement(C,B)}
    \end{array}
    \right\}
    \end{array}
\]
From this hypothesis, \name{} identifies the MUSP:
\[
    \begin{array}{l}
    \left\{
    \begin{array}{l}
        \emph{$\leftarrow$ zero(C), decrement(C,B)}
    \end{array}
    \right\}
    \end{array}
\]
\noindent
This program is unsatisfiable because the BK for this task is defined for natural numbers, so zero has no predecessor.
\name{} prunes the specialisations and redundancies of this MUSP, thus vastly reducing the hypothesis space.

Likewise, on the \emph{buttons} task, \name{} sometimes generates the hypothesis:
\[
    \begin{array}{l}
    \left\{
    \begin{array}{l}
        \emph{next(A,B) $\leftarrow$ my\_true(A,B), succ(B,C), succ(C,B)}
    \end{array}
    \right\}
    \end{array}
\]
From this hypothesis, \name{} identifies the MUSP:
\[
    \begin{array}{l}
    \left\{
    \begin{array}{l}
        \emph{$\leftarrow$ 
    succ(B,C), succ(C,B)}
    \end{array}
    \right\}
    \end{array}
\]
\noindent
This program is unsatisfiable because the successor relation (\emph{succ}) is asymmetric. 
\name{} prunes any hypothesis that contains this pair of literals.

Similarly, on the \emph{length} task, \name{} identifies the MUSP:
\[
    \begin{array}{l}
    \left\{
    \begin{array}{l}
        \emph{
        f(A,B)$ \leftarrow$ one(B), tail(A,C), tail(C,D)}
    \end{array}
    \right\}
    \end{array}
\]
\noindent
This program is unsatisfiable because if a list $A$ has length $B=1$ then its tail $C$ does not have a tail $D$.

\name{} reduces learning times when it discovers small MUSPs early in the search but is otherwise less helpful.
For instance, \name{} does not help much on the \emph{zendo} tasks because these tasks have many background relations that almost always hold, such as \emph{has\_piece(State, Piece)}, because every Zendo game state has a piece, and \emph{size(Piece, Size)}, because a piece always has a size. 

Overall, the results in this section show that the answer to \textbf{Q1} is yes, identifying MUSPs can drastically improve learning times whilst maintaining high predictive accuracies. 



\begin{table}[ht!]
\centering
\small
\begin{tabular}{@{}l|ccc@{}}
    \textbf{Domain} & \textbf{\popper{}} & \textbf{\name{}} & \textbf{Change}\\
    \midrule
\emph{trains} & 1630 $\pm$ 160 & 1632 $\pm$ 159 & +0\% \\
\emph{zendo} & 2303 $\pm$ 264 & 2304 $\pm$ 261 & +0\% \\
\emph{imdb} & 177 $\pm$ 40 & 162 $\pm$ 36 & \textbf{--8\%} \\
\emph{iggp} & 43122 $\pm$ 6189 & 16623 $\pm$ 2795 & \textbf{--61\%} \\
\emph{krk} & 440 $\pm$ 71 & 409 $\pm$ 60 & \textbf{--7\%} \\
\emph{synthesis} & 4395 $\pm$ 387 & 1595 $\pm$ 162 & \textbf{--63\%} \\
\emph{sql} & 14328 $\pm$ 1932 & 44 $\pm$ 4 & \textbf{--99\%} \\
\end{tabular}
\caption{
Number of programs considered.
}
\label{tab:num_progs}
\end{table}

\begin{table}[ht!]
\small
\centering
\begin{tabular}{@{}l|ccc@{}}
    \textbf{Domain} & \textbf{Total} & \textbf{MUSPs} & \textbf{Ratio}\\
    \midrule
\emph{trains} & 13 $\pm$ 1 & 0 $\pm$ 0 & 0\% \\
\emph{zendo} & 36 $\pm$ 4 & 0.1 $\pm$ 0 & 0\% \\
\emph{imdb} & 148 $\pm$ 39 & 3 $\pm$ 1 & 2\% \\
\emph{iggp} & 207 $\pm$ 34 & 4 $\pm$ 0 & 1\% \\
\emph{krk} & 38 $\pm$ 5 & 0.7 $\pm$ 0.1 & 1\% \\
\emph{synthesis} & 199 $\pm$ 33 & 4 $\pm$ 1 & 2\% \\
\emph{sql} & 13 $\pm$ 1 & 3 $\pm$ 0 & 23\% \\
\end{tabular}
\caption{
The proportion of the learning time (seconds) spent finding MUSPs.
}
\label{tab:overhead}
\end{table}

\subsubsection{Q2. How Does \name{} Compare to Other Approaches?}

Table \ref{tab:qaccs} shows the predictive accuracies aggregated per domain of all the systems.
\name{} has higher predictive accuracy than \metagol{} on every domain.
\name{} has equal or higher predictive accuracy than \ale{}{} on all but two domains: \emph{trains}, where the difference is negligible, and \emph{krk}.

The most comparable approach is \disco{}.
The results in Table \ref{tab:qaccs} show that \name{} has equal or higher predictive accuracy than \disco{} on all the domains. 
Table \ref{tab:sota_times} shows the learning times aggregated per domain for \name{} and \disco{}.
The results show that \name{} outperforms \disco{} in 6/7 domains and that they are matched in the other domain.
A paired t-test confirms the significance of the difference ($p < 0.01$). 
\name{} can reduce learning times by 97\% compared to \disco{}, such as on the \emph{sql} domain.

The two main reasons for the performance difference between \name{} and \disco{} are that \disco{} (i) only finds unsatisfiable rules from the BK alone and does not consider examples, and (ii) uses a predefined set of properties to find unsatisfiable rules.
For instance, for the \emph{sql-02} task, we want to learn an \emph{out/2} relation using BK with \emph{family/4} facts.
For this task, \disco{} does not learn any constraints from the BK.
By contrast, \name{} quickly learns many MUSPs, including: 
\[
    \begin{array}{l}
    \left\{
    \begin{array}{l}
        \emph{out(A,B) $\leftarrow$ family(C,A,D,E)}     \end{array}
    \right\}
    \end{array}
\]
\noindent
This rule is a MUSP because there is no positive example $out(A,B)$ where $A$ is true because the second argument in a \emph{family/4} fact in the BK.
Because it only considers the BK and not the examples, \textsc{disco} cannot learn this MUSP.
\name{} also learns the MUSP:
\[
    \begin{array}{l}
    \left\{
    \begin{array}{l}
        \emph{$\leftarrow$ family(D,B,C,E), family(C,D,E,A)}\\
    \end{array}
    \right\}
    \end{array}
\]
\textsc{disco} cannot learn this MUSP because it is not one of its predefined properties.
On this task, \disco{} takes 1005s to terminate.
By contrast, \name{} learns an optimal solution and terminates in only 9s, a 97\% improvement.

Overall, these results suggest that the answer to \textbf{Q2} is that \name{} compares favourably to existing approaches and can substantially improve learning performance, both in terms of predictive accuracies and learning times.

\begin{table}[ht!]
\centering
\begin{tabular}{@{}l|ccc@{}}
\textbf{Domain} & 
\textbf{\textsc{Disco}} & 
\textbf{\name{}} & 
\textbf{Change} \\

\midrule
\emph{trains} & 14 $\pm$ 1 & 13 $\pm$ 1 & \textbf{--7\%} \\
\emph{zendo} & 40 $\pm$ 5 & 36 $\pm$ 4 & \textbf{--10\%} \\
\emph{imdb} & 250 $\pm$ 73 & 148 $\pm$ 39 & \textbf{--40\%} \\
\emph{iggp} & 583 $\pm$ 66 & 207 $\pm$ 34 & \textbf{--64\%} \\
\emph{krk} & 1219 $\pm$ 153 & 1214 $\pm$ 154 & 0\% \\
\emph{synthesis} & 327 $\pm$ 38 & 199 $\pm$ 33 & \textbf{--39\%} \\
\emph{sql} & 505 $\pm$ 58 & 13 $\pm$ 1 & \textbf{--97\%} \\

\end{tabular}
\caption{
Learning times (seconds).
}
\label{tab:sota_times}
\end{table}

\section{Conclusions and Limitations}
We have introduced an approach that identifies \emph{minimal unsatisfiable subprograms} (MUSPs).
We have shown that pruning with MUSPs is optimally sound (Propositions \ref{prop_spec_sound} and \ref{prop_redund_sound}) and effective (Propositions \ref{prop_spec} and \ref{prop_redund}).
We implemented our approach in \name{}, which identifies the MUSPs of definite programs, including recursive programs.
Our experimental results on many diverse domains show that our approach can drastically reduce learning times whilst maintaining high predictive accuracy, sometimes reducing learning times by 99\%.
In addition to the substantial empirical improvements, we think that a key contribution of this paper is to show that a well-studied idea in constraint satisfaction (finding minimal unsatisfiable cores) can drastically improve the performance of a machine learning algorithm.


\paragraph{Limitations.}
To find MUSPs, we use what Dershowitz et al. \shortcite{DBLP:conf/sat/DershowitzHN06} call a \emph{folk algorithm} where we delete literals one by one.
As Dershowitz et al. \shortcite{DBLP:conf/sat/DershowitzHN06} point out, this folk algorithm does not scale well to large formulas.
Although scalability is not an issue in our experiments, future work should address this limitation.



\bibliographystyle{named}
\bibliography{ourbib15}

 \begin{appendices}
 
\section{Terminology}
\label{sec:bk}
\subsection{Logic Programming}
We assume familiarity with logic programming \cite{lloyd:book} but restate some key relevant notation. A \emph{variable} is a string of characters starting with an uppercase letter. A \emph{predicate} symbol is a string of characters starting with a lowercase letter. The \emph{arity} $n$ of a function or predicate symbol is the number of arguments it takes. An \emph{atom} is a tuple $p(t_1, ..., t_n)$, where $p$ is a predicate of arity $n$ and $t_1$, ..., $t_n$ are terms, either variables or constants. An atom is \emph{ground} if it contains no variables. A \emph{literal} is an atom or the negation of an atom. A \emph{clause} is a set of literals.
A \emph{clausal theory} is a set of clauses. A \emph{constraint} is a clause without a positive literal. A \emph{definite} clause is a clause with exactly one positive literal. A \emph{program} is a set of definite clauses. A \emph{substitution} $\theta = \{v_1 / t_1, ..., v_n/t_n \}$ is the simultaneous replacement of each variable $v_i$ by its corresponding term $t_i$. 
A clause $c_1$ \emph{subsumes} a clause $c_2$ if and only if there exists a substitution $\theta$ such that $c_1 \theta \subseteq c_2$. 
A program $h_1$ subsumes a program $h_2$, denoted $h_1 \preceq h_2$, if and only if $\forall c_2 \in h_2, \exists c_1 \in h_1$ such that $c_1$ subsumes $c_2$. A program $h_1$ is a \emph{specialisation} of a program $h_2$ if and only if $h_2 \preceq h_1$. A program $h_1$ is a \emph{generalisation} of a program $h_2$ if and only if $h_1 \preceq h_2$.
\subsection{Answer Set Programming}
We also assume familiarity with answer set programming \cite{asp} but restate some key relevant notation \cite{ilasp}.
A \emph{literal} can be either an atom $p$ or its \emph{default negation} $\text{not } p$ (often called \emph{negation by failure}). A normal rule is of the form $h \leftarrow b_1, ..., b_n, \text{not } c_1,... \text{not } c_m$. where $h$ is the \emph{head} of the rule, $b_1, ..., b_n, \text{not } c_1,... \text{not } c_m$ (collectively) is the \emph{body} of the rule, and all $h$, $b_i$, and $c_j$ are atoms. A \emph{constraint} is of the form $\leftarrow b_1, ..., b_n, \text{not } c_1,... \text{not } c_m.$ where the empty head means false. A \emph{choice rule} is an expression of the form $l\{h_1,...,h_m\}u \leftarrow b_1,...,b_n, \text{not } c_1,... \text{not } c_m$ where the head $l\{h_1,...,h_m\}u$ is called an \emph{aggregate}. In an aggregate, $l$ and $u$ are integers and $h_i$, for $1 \leq i \leq m$, are atoms. An \emph{answer set program} $P$ is a finite set of normal rules, constraints and choice rules. Given an answer set program $P$, the \emph{Herbrand base} of $P$, denoted
as ${HB}_P$, is the set of all ground (variable free) atoms that can be formed from the predicates and constants that appear in $P$. When $P$ includes only normal rules, a set $A \in {HB}_P$ is an \emph{answer set} of $P$ iff it is the minimal model of the  \emph{reduct} $P^A$, which is the program constructed from the grounding of $P$ by first removing any rule whose body contains a literal $\text{not } c_i$ where $c_i \in A$, and then removing any defaultly negated literals in the remaining rules. An answer set $A$ satisfies a ground constraint $\leftarrow b_1, ..., b_n, \text{not } c_1,... \text{not } c_m.$ if it is not the case that $\{b_1, ..., b_n\} \in A$ and $A \cap \{c_1, ..., c_m\} = \emptyset$.
\section{Redundant Programs}
\begin{example}[\textbf{Redundant hypothesis}] 
Assume that the following hypothesis is totally incomplete:
\[
    \begin{array}{l}
    h_1 = \left\{
    \begin{array}{l}
    \emph{f(A,B) $\leftarrow$ head(A,B)}
    \end{array}
    \right\}
    \end{array}
\]
Then the following hypothesis is redundant with respect to $h_1$:
\[
    \begin{array}{l}
    \left\{
    \begin{array}{l}
    \emph{f(A,B) $\leftarrow$ head(A,B), one(B)}\\
    \emph{f(A,B) $\leftarrow$ tail(A,B), empty(B)}
    \end{array}
    \right\}
    \end{array}
\]
The following hypothesis is not redundant with respect to $h_1$ because its second rule is recursive and does not specialise a rule in $h_1$:
\[
    \begin{array}{l}
    \left\{
    \begin{array}{l}
    \emph{f(A,B) $\leftarrow$ head(A,B), one(B)}\\
    \emph{f(A,B) $\leftarrow$ tail(A,C), f(C,B)}
    \end{array}
    \right\}
    \end{array}
\]
By contrast, the following hypothesis is redundant with respect to $h_1$ because, although the second rule is recursive, it specialises the rule in $h_1$:
\[
    \begin{array}{l}
    \left\{
    \begin{array}{l}
    \emph{f(A,B) $\leftarrow$ head(A,B), one(B)}\\
    \emph{f(A,B) $\leftarrow$ head(A,B), tail(A,C), f(C,B)}
    \end{array}
    \right\}
    \end{array}
\]
\end{example}

\section{\name{} Correctness}
\label{sec:proof}
We show the correctness of \name{}.
To show this result, we first show that the constraints introduced by \name{} never prune optimal solutions from the hypothesis space.
In contrast to \popper{}, \name{} additionally prunes (a) specialisations and (b) redundancies of MUSPs.
We show that specialisation constraints for MUSPs are sound i.e. they do not prune solutions from the hypothesis space (Proposition \ref{specialisation}) and that redundancy constraints for MUSPs are optimally sound i.e. they do not prune optimal solutions from the hypothesis space (Proposition \ref{redundancy}).

We show a result for case (a):
\begin{proposition}[\textbf{MUSP specialisations}]
Let $h$ be a program, $m$ be a MUSP of $h$, and $h'$ be a specialisation of $m$.
Then $h'$ is not a solution.
\label{specialisation}
\end{proposition}
\begin{proof}
Since $m$ is totally incomplete and $h'$ is a specialisation of $m$ then $h'$ is totally incomplete. Therefore $h'$ is not a solution.
\end{proof}

\noindent
We show a result for case (b). 

\noindent
\begin{proposition}[\textbf{MUSP redundancy}]
Let $h$ and $h'$ be programs, $m$ be a MUSPs of $h$, and $h'$ be redundant with respect to $m$.
Then $h'$ is not an optimal solution.
\label{redundancy}
\end{proposition}
\begin{proof}
Since $m$ is a MUSP then it is totally incomplete. 
Cropper and Morel \shortcite{poppi} show that hypotheses redundant with respect to a totally incomplete hypothesis are not optimal solutions (Theorem 1). 
Therefore $h'$ is not an optimal solution.
\end{proof}

\noindent
We now show that using MUSPs to prune redundant hypotheses leads to more pruning:

\begin{proposition}[\textbf{Specialisation pruning}] 
Let $h$ be a hypothesis, $m$ be a MUSP of $h$, $S_h$ be the specialisations of $h$, and $S_m$ be the specialisations of $m$.
Then $S_h \subset S_m$.
\end{proposition}
\begin{proof}
Since $h$ specialises $m$ and $size(h) > size(m)$ then (i) $S_m$ includes $h$ and $S_h$, and (ii) $S_h$ does not include $m$.
\end{proof}

\begin{proposition}[\textbf{Redundancy pruning}] 
Let $h$ be a hypothesis,
$m$ be a MUSP of $h$, 
$S_h$ be the hypotheses redundant with respect to $h$,
and $S_m$ be the hypotheses redundant with respect to $m$.
Then $S_h \subset S_m$.
\end{proposition}
\begin{proof}
We show that $h' \in S_h$  implies $h' \in S_m$.
Since $h'$ is redundant with respect to $h$ then there is no recursive rule $r'$ in $h'$ that does not specialise a rule in $h$.
We show that there is no recursive rule $r'$ in $h'$ that does not specialise a rule in $m$.
For contradiction, suppose there is such a rule $r'$, which implies that no rule in $m$ generalises $r'$.
By condition (ii) in the definition of a core (Definition 4), $m$ subsumes $h$.
Since $m$ subsumes $h$ then every rule in $h$ is subsumed by a rule in $m$, including $r'$, which leads to a contradiction. 
Therefore $h' \in S_m$.
Moreover, $m \in S_m$ but $m \not\in S_h$ since $h$ specialises $m$ and $size(m) < size(h)$. 
Therefore $S_h \subset S_m$, which completes the proof.

\end{proof}



\noindent
We use the Propositions \ref{specialisation} and \ref{redundancy} to prove the correctness \name{}.

\begin{theorem}[\textbf{Correctness}]
\name{} returns an optimal solution if one exists.
\end{theorem}
\begin{proof}
Cropper and Morel \shortcite{popper} show that given optimally sound constraints then \popper{} returns an optimal solution if one exists (Theorem 1).
\name{} builds on \popper{} by additionally pruning specialisations and redundancies of MUSPs, which are optimally sound by Propositions \ref{specialisation} and \ref{redundancy} respectively.
Therefore \name{} never prunes optimal solutions so returns an optimal solution if one exists.
\end{proof}

\section{Experiments}
\label{sec:domains}

\subsection{Experimental domains}
We describe the characteristics of the domains and tasks used in our experiments in Tables \ref{tab:dataset} and \ref{tab:tasks}. Figure \ref{fig:sols} shows example solutions for some of the tasks.

\begin{table}[ht!]
\footnotesize
\centering
\begin{tabular}{@{}l|cccc@{}}
\textbf{Task} & \textbf{\# examples} & \textbf{\# relations} & \textbf{\# constants} & \textbf{\# facts}\\
\midrule
\emph{trains} & 1000 & 20 & 8561 & 28503 \\
\midrule
\emph{zendo1} & 100 & 16 & 1049 & 2270\\
\emph{zendo2} & 100 & 16 & 1047 & 2184\\
\emph{zendo3} & 100 & 16 & 1100 & 2320\\
\emph{zendo4} & 100 & 16 & 987 & 2087\\
\midrule
\emph{imdb1} & 383 & 6 & 299 & 1330 \\
\emph{imdb2} & 71825 & 6 & 299 & 1330 \\
\emph{imdb3} & 121801 & 6 & 299 & 1330 \\
\midrule
\emph{krk1} & 20 & 10 & 108 & 4223 \\
\emph{krk2} & 20 & 10 & 108 & 168 \\
\emph{krk3} & 20 & 8 & 108 & 140 \\
\midrule
\emph{md} & 54 & 12 & 13 & 29 \\
\emph{buttons} & 530 & 13 & 60 & 656 \\
\emph{buttons-g} & 44 & 21 & 31 & 130 \\
\emph{rps} & 464 & 6 & 64 & 405 \\
\emph{coins} & 2544 & 9 & 110 & 1101 \\
\emph{coins-g} & 125 & 118 & 68 & 742 \\
\emph{attrition} & 672 & 12 & 65 & 163\\
\emph{centipede} & 26 & 34 & 61 & 138 
\\\midrule
\emph{contains} & 20 & 10 & $\infty$ & $\infty$\\
\emph{dropk} & 20 & 10 & $\infty$ & $\infty$\\
\emph{droplast} & 20 & 10 & $\infty$ & $\infty$\\
\emph{evens} & 20 & 10 & $\infty$ & $\infty$\\
\emph{finddup} & 20 & 10 & $\infty$ & $\infty$\\
\emph{last} & 20 & 10 & $\infty$ & $\infty$\\
\emph{len} & 20 & 10 & $\infty$ & $\infty$\\
\emph{reverse} & 20 & 10 & $\infty$ & $\infty$\\
\emph{sorted} & 20 & 10 & $\infty$ & $\infty$\\
\emph{sumlist} & 20 & 10 & $\infty$ & $\infty$\\
\emph{next} & 20 & 10 & $\infty$ & $\infty$\\
\emph{rotateN} & 20 & 12 & $\infty$ & $\infty$\\
\emph{inttobin} & 20 & 13 & $\infty$ & $\infty$\\
\emph{chartointodd} & 20 & 12 & $\infty$ & $\infty$\\
\emph{twosucc} & 20 & 10 & $\infty$ & $\infty$
\\\midrule
\emph{sql-01} & 2 & 4 & 15 & 22\\
\emph{sql-02} & 1 & 2 & 10 & 3\\
\emph{sql-03} & 2 & 2 & 8 & 4\\
\emph{sql-04} & 6 & 3 & 13 & 9\\
\emph{sql-05} & 5 & 3 & 17 & 12\\
\emph{sql-06} & 9 & 3 & 18 & 9\\
\emph{sql-07} & 5 & 2 & 13 & 5\\
\emph{sql-08} & 2 & 4 & 8 & 6\\
\emph{sql-09} & 1 & 4 & 9 & 6\\
\emph{sql-10} & 2 & 3 & 12 & 10\\
\emph{sql-11} & 2 & 7 & 31 & 30\\
\emph{sql-12} & 7 & 6 & 30 & 36\\
\emph{sql-13} & 7 & 3 & 18 & 17\\
\emph{sql-14} & 6 & 4 & 23 & 11\\
\emph{sql-15} & 7 & 4 & 23 & 50\\
\end{tabular}
\caption{
Experimental domain description.
}

\label{tab:dataset}
\end{table}

\begin{table}[ht!]
\footnotesize
\centering
\begin{tabular}{@{}l|ccp{5mm}c@{}}
\textbf{Task} & \textbf{\#rules} & \textbf{\#literals} & \textbf{max rule size} & \textbf{recursion}\\
\midrule
\emph{train1} & 1 & 6 &  6 & no\\
\emph{train2} & 2 & 11 & 6 & no\\
\emph{train3} & 3 & 17 &  7& no\\
\emph{train4} & 4 & 26 &  7& no\\
\midrule
\emph{zendo1} & 1 & 7 & 7 & no\\
\emph{zendo2} & 2 & 14 & 7 & no\\
\emph{zendo3} & 3 & 20 & 7& no\\
\emph{zendo4} & 4 & 23 & 7& no\\
\midrule
\emph{imdb1} & 1 & 5 & 5 & no\\
\emph{imdb2} & 1 & 5 & 5 & no\\
\emph{imdb3} & 2 & 10 & 5 & no\\
\midrule
\emph{krk1} & 1 & 8 & 8 & no\\
\emph{krk2} & 1 & 36 & 9 & no\\
\emph{krk3} & 2 & 16 & 8 & no\\
\midrule
\emph{md} & 2 & 11 & 6 & no\\
\emph{buttons} & 10 & 61 & 7& no\\
\emph{rps} & 4 & 25 & 7& no\\
\emph{coins} & 16 & 45 & 7& no\\
\emph{buttons-g} & 4 & 24 & 9 & no\\
\emph{coins-g} & 4 & 21 & 6 & no \\
\emph{attrition} & 3 & 14 & 5& no\\
\emph{centipede} & 2 & 8 & 4& no\\
\midrule
\emph{contains} & 3 & 9 & 3& yes\\
\emph{dropk} & 2 & 7 & 4& yes\\
\emph{droplast} & 2 & 8 & 5& yes\\
\emph{evens} & 2 & 7 & 5& yes\\
\emph{finddup} & 2 & 7 & 4& yes\\
\emph{last} & 2 & 7 & 4& yes\\
\emph{len} & 2 & 7 & 4& yes\\
\emph{reverse} & 2 & 8 & 5& yes\\
\emph{sorted} & 2 & 9 & 6& yes\\
\emph{sumlist} & 2 & 7 & 5& yes\\
\emph{next} & 2 & 7 & 4& yes\\
\emph{rotateN} & 2 & 9 & 6& yes\\
\emph{inttobin} & 2 & 10 & 7& yes\\
\emph{chartointodd} & 2 & 8 & 6& yes\\
\emph{twosucc} & 2 & 8 & 5& yes\\
\midrule
\emph{sql-01} & 1 & 4 & 4& no\\
\emph{sql-02} & 1 & 4 & 4& no\\
\emph{sql-03} & 1 & 4 & 4& no\\
\emph{sql-04} & 2 & 8 & 4& no\\
\emph{sql-05} & 1 & 5 & 5& no\\
\emph{sql-06} & 2 & 5 & 3& no\\
\emph{sql-07} & 1 & 4 & 4& no\\
\emph{sql-08} & 1 & 5 & 5& no\\
\emph{sql-09} & 1 & 2 & 2& no\\
\emph{sql-10} & 1 & 3 & 3& no\\
\emph{sql-11} & 1 & 5 & 5& no\\
\emph{sql-12} & 1 & 5 & 5& no\\
\emph{sql-13} & 1 & 3 & 3& no\\
\emph{sql-14} & 1 & 2 & 2& no\\
\emph{sql-15} & 1 & 4 & 4& no\\
\end{tabular}
\caption{
Statistics about the optimal solutions for each task.
For instance, the optimal solution for \emph{buttons} has 10 rules and 61 literals and the largest rule has 7 literals.
}

\label{tab:tasks}
\end{table}

\paragraph{Michalski trains.}
The goal of these tasks is to find a hypothesis that distinguishes eastbound and westbound trains \cite{michalski:trains}. There are four increasingly complex tasks. 
There are 1000 examples but the distribution of positive and negative examples is different for each task.
We randomly sample the examples and split them into 80/20 train/test partitions.

\paragraph{Zendo.} Zendo is an inductive game in which one player, the Master, creates a rule for structures made of pieces with varying attributes to follow. The other players, the Students, try to discover the rule by building and studying structures which are labelled by the Master as following or breaking the rule. The first student to correctly state the rule wins. We learn four increasingly complex rules for structures made of at most 5 pieces of varying color, size, orientation and position. 
Zendo is a challenging game that has attracted much interest in cognitive science \cite{zendo}.

\paragraph{IMDB.}
The real-world IMDB dataset \cite{imdb} includes relations between movies, actors, directors, movie genre, and gender. It has been created from the International Movie Database (IMDB.com) database. We learn the relation \emph{workedunder/2}, a more complex variant \emph{workedwithsamegender/2}, and the disjunction of the two.
This dataset is frequently used \cite{alps}.

\paragraph{Chess.} The task is to learn chess patterns in the king-rook-king (\emph{krk}) endgame, which is the chess ending with white having a king and a rook and black having a king. We learn the concept of white rook protection by the white king (\emph{krk1}) \cite{celine:bottom}, king opposition (\emph{krk2}), and rook attack (\emph{krk3}).

\paragraph{IGGP.}
In \emph{inductive general game playing} (IGGP)  \cite{iggp} the task is to induce a hypothesis to explain game traces from the general game playing competition \cite{ggp}.
IGGP is notoriously difficult for machine learning approaches.
The currently best-performing system can only learn perfect solutions for 40\% of the tasks.
Moreover, although seemingly a toy problem, IGGP is representative of many real-world problems, such as inducing semantics of programming languages \cite{DBLP:conf/ilp/BarthaC19}. 
We use six games: \emph{minimal decay (md)}, \emph{buttons}, \emph{rock - paper - scissors (rps)}, \emph{coins}, \emph{attrition}, and \emph{centipede}.



\paragraph{Program Synthesis.} This dataset includes list transformation tasks. It involves learning recursive programs which has been identified as a difficult challenge for ILP systems \cite{ilp20}. We add 5 new tasks.

\paragraph{SQL.} This dataset \cite{wang2017synthesizing,prosynth} comprises 15  real-world tasks taken from online forums. These tasks involve synthesising SQL queries that can be expressed in Datalog. Since there are no testing data, we report training accuracy. This dataset contains only positive examples.


\subsection{Experimental Setup}
We measure the mean and standard error of the predictive accuracy and learning time.
We use a m5a AWS instance with 64vCPU and 256GB of memory. 
All the systems use a single CPU.



\subsubsection{Comparison against other ILP systems}

To answer \textbf{Q2} we compare \name{} against \popper{}, \ale{}, \metagol{}, and \disco{}, which we describe below.

\begin{description}
\item[\popper{}] \popper{} uses identical biases to \name{} so the comparison is direct, i.e. fair.
We use \popper{} 2.0.0 \cite{combo}.
\item[\ale{}] \ale{} excels at learning many large non-recursive rules and should excel at the trains and IGGP tasks.
Although \ale{} can learn recursive programs, it struggles to do so.
\name{} and \ale{} use similar biases so the comparison can be considered reasonably fair.
For instance, Figures \ref{fig:biasname}  and \ref{fig:biasale} show the \name{} and \ale{} biases used for the Zendo tasks.

\begin{figure}[ht!]
\begin{lstlisting}[caption=\name{}]
max_clause(6).
max_vars(6).
max_body(6).

head_pred(zendo,1).
body_pred(piece,2).
body_pred(contact,2).
body_pred(coord1,2).
body_pred(coord2,2).
body_pred(size,2).
body_pred(blue,1).
body_pred(green,1).
body_pred(red,1).
body_pred(small,1).
body_pred(medium,1).
body_pred(large,1).
body_pred(upright,1).
body_pred(lhs,1).
body_pred(rhs,1).
body_pred(strange,1).

type(zendo,(state,)).
type(piece,(state,piece)).
type(contact,(piece,piece)).
type(coord1,(piece,real)).
type(coord2,(piece,real)).
type(size,(piece,real)).
type(blue,(piece,)).
type(green,(piece,)).
type(red,(piece,)).
type(small,(real,)).
type(medium,(real,)).
type(large,(real,)).
type(upright,(piece,)).
type(lhs,(piece,)).
type(rhs,(piece,)).
type(strange,(piece,)).

direction(zendo,(in,)).
direction(piece,(in,out)).
direction(contact,(in,out)).
direction(coord1,(in,out)).
direction(coord2,(in,out)).
direction(size,(in,out)).
direction(blue,(in,)).
direction(green,(in,)).
direction(red,(in,)).
direction(small,(in,)).
direction(medium,(in,)).
direction(large,(in,)).
direction(upright,(in,)).
direction(lhs,(in,)).
direction(rhs,(in,)).
direction(strange,(in,)).
\end{lstlisting}
\caption{The bias files for \name{} for the learning tasks \emph{zendo}.}
\label{fig:biasname}
\end{figure}

\begin{figure}
\begin{lstlisting}[caption=\ale{}]
:- aleph_set(i,6).
:- aleph_set(clauselength,7).
:- aleph_set(nodes,50000).

:- modeh(*,zendo(+state)).
:- modeb(*,piece(+state,-piece)).
:- modeb(*,contact(+piece,-piece)).
:- modeb(*,coord1(+piece,-real)).
:- modeb(*,coord2(+piece,-real)).
:- modeb(*,size(+piece,-real)).
:- modeb(*,blue(+piece)).
:- modeb(*,green(+piece)).
:- modeb(*,red(+piece)).
:- modeb(*,small(+real)).
:- modeb(*,medium(+real)).
:- modeb(*,large(+real)).
:- modeb(*,upright(+piece)).
:- modeb(*,lhs(+piece)).
:- modeb(*,rhs(+piece)).
:- modeb(*,strange(+piece)).

\end{lstlisting}
\caption{The bias files for \ale{} for the learning tasks \emph{zendo}.}
\label{fig:biasale}
\end{figure}

\item[\metagol{}] \metagol{} is one of the few systems that can learn recursive Prolog programs.
\metagol{} uses user-provided \emph{metarules} (program templates) to guide the search for a solution.
We use the approximate universal set of metarules described by Cropper and Tourret \shortcite{reduce}.
However, these metarules are only sufficient to learn programs with literals of arity at most two, and so \metagol{} cannot solve many tasks we consider.

\item[\disco{}]
\disco{} discovers constraints from the BK as a preprocessing step.
It uses a predefined set of constraints, such as \emph{asymmetry} and \emph{antitransitivity}.
It uses these constraints to bootstrap \popper{}.

\end{description}
\noindent
We also tried/considered other ILP systems.
We considered \ilasp{} \cite{ilasp}.
However, \ilasp{} builds on \aspal{} and first precomputes every possible rule in a hypothesis space, which is infeasible for our datasets.
For instance, it would require precomputing $10^{15}$ rules for the \emph{coins} task. 
In addition, \ilasp{} cannot learn Prolog programs so is unusable in the synthesis tasks.
We also considered \textsc{hexmil} \cite{hexmil} and \textsc{louise} \cite{louise}, both metarule-based approaches similar to \metagol{}.
However, their performance was considerably worse than \metagol{} so we have excluded them from the comparison.

\begin{figure*}
\centering
\footnotesize
\begin{lstlisting}[caption=trains2\label{trains2}]
east(A):-car(A,C),roof_open(C),load(C,B),triangle(B)
east(A):-car(A,C),car(A,B),roof_closed(B),two_wheels(C),roof_open(C).
\end{lstlisting}

\centering
\footnotesize
\begin{lstlisting}[caption=trains4\label{trains4}]
east(A):-has_car(A,D),has_load(D,B),has_load(D,C),rectangle(B),diamond(C).
east(A):-has_car(A,B),has_load(B,C),hexagon(C),roof_open(B),three_load(C).
east(A):-has_car(A,E),has_car(A,D),has_load(D,C),triangle(C),has_load(E,B),hexagon(B).
east(A):-has_car(A,C),roof_open(C),has_car(A,B),roof_flat(B),short(C),long(B).
\end{lstlisting}
\centering

\begin{lstlisting}[caption=zendo1\label{zendo1}]
zendo1(A):- piece(A,C),size(C,B),blue(C),small(B),contact(C,D),red(D).
\end{lstlisting}
\centering

\begin{lstlisting}[caption=zendo2\label{zendo2}]
zendo2(A):- piece(A,B),piece(A,D),piece(A,C),green(D),red(B),blue(C).
zendo2(A):- piece(A,D),piece(A,B),coord1(B,C),green(D),lhs(B),coord1(D,C).
\end{lstlisting}
\centering

\begin{lstlisting}[caption=zendo3\label{zendo3}]
zendo3(A):- piece(A,D),blue(D),coord1(D,B),piece(A,C),coord1(C,B),red(C).
zendo3(A):- piece(A,D),contact(D,C),rhs(D),size(C,B),large(B).
zendo3(A):- piece(A,B),upright(B),contact(B,D),blue(D),size(D,C),large(C).
\end{lstlisting}
\centering

\begin{lstlisting}[caption=zendo4\label{zendo4}]
zendo4(A):- piece(A,C),contact(C,B),strange(B),upright(C).
zendo4(A):- piece(A,D),contact(D,C),coord2(C,B),coord2(D,B).
zendo4(A):- piece(A,D),contact(D,C),size(C,B),red(D),medium(B).
zendo4(A):- piece(A,D),blue(D),lhs(D),piece(A,C),size(C,B),small(B).\end{lstlisting}
\centering

\begin{lstlisting}[caption=krk1\label{krk1}]
f(A):-cell(A,B,C,D),white(C),cell(A,E,C,F),rook(D),king(F),distance(B,E,G),one(G).
\end{lstlisting}

\begin{lstlisting}[caption=krk2\label{krk2}]
f(A):-cell(A,B,C,D),white(C),cell(A,E,F,D),king(D),black(F),samerank(B,E),
      nextfile(B,G),nextfile(G,E).
f(A):-cell(A,B,C,D),white(C),cell(A,E,F,D),king(D),black(F),samerank(B,E),
      nextfile(E,G),nextfile(G,B).
f(A):-cell(A,B,C,D),white(C),cell(A,E,F,D),king(D),black(F),samefile(B,E),
      nextrank(B,G),nextrank(G,E).
f(A):-cell(A,B,C,D),white(C),cell(A,E,F,D),king(D),black(F),samefile(B,E),
      nextrank(F,G),nextrank(G,B).
\end{lstlisting}

\begin{lstlisting}[caption=krk3\label{krk3}]
f(A):-cell(A,B,C,D),white(C),cell(A,E,F,G),rook(D),king(G),black(F),samerank(B,E).
f(A):-cell(A,B,C,D),white(C),cell(A,E,F,G),rook(D),king(G),black(F),samefile(B,E).
\end{lstlisting}

\begin{lstlisting}[caption=minimal decay\label{minimaldecay}]
next_value(A,B):-c_player(D),c_pressButton(C),c5(B),does(A,D,C).
next_value(A,B):-c_player(C),my_true_value(A,E),does(A,C,D),my_succ(B,E),c_noop(D).
\end{lstlisting}

\begin{lstlisting}[caption=rps\label{rps}]
next_score(A,B,C):-does(A,B,E),different(G,B),my_true_score(A,B,F),beats(E,D),
                   my_succ(F,C),does(A,G,D).
next_score(A,B,C):-different(G,B),beats(D,F),my_true_score(A,E,C),does(A,G,D),does(A,E,F).
next_score(A,B,C):-my_true_score(A,B,C),does(A,B,D),does(A,E,D),different(E,B).
\end{lstlisting}


\caption{Example solutions.}
\end{figure*}

\setcounter{figure}{3}






\begin{figure*}
\centering
\begin{lstlisting}[caption=contains]
contains(A):- head(A,B),c_6(B).
contains(A):- head(A,B),c_9(B).
contains(A):- tail(A,B),contains(B).
\end{lstlisting}

\centering
\begin{lstlisting}[caption=dropk]
dropk(A,B,C):- tail(A,C),odd(B),one(B).
dropk(A,B,C):- decrement(B,E),tail(A,D),dropk(D,E,C).
\end{lstlisting}


\centering
\begin{lstlisting}[caption=evens]
evens(A):- empty(A).
evens(A):- head(A,B),even(B),tail(A,C),evens(C).
\end{lstlisting}

\centering
\begin{lstlisting}[caption=finddup]
finddup(A,B):- head(A,B),tail(A,C),element(C,B).
finddup(A,B):- tail(A,C),finddup(C,B).
\end{lstlisting}

\centering
\begin{lstlisting}[caption=last]
last(A,B):- head(A,B),tail(A,C),empty(C).
last(A,B):- tail(A,C),last(C,B).
\end{lstlisting}

\centering
\begin{lstlisting}[caption=reverse]
reverse(A,B):- empty_out(B),empty(A).
reverse(A,B):- head(A,D),tail(A,E),reverse(E,C),append(C,D,B).
\end{lstlisting}

\centering
\begin{lstlisting}[caption=sorted]
sorted(A):-tail(A,B),empty(B).
sorted(A):-tail(A,D),head(A,B),head(D,C),geq(C,B),sorted(D).
\end{lstlisting}

\centering
\begin{lstlisting}[caption=sumlist]
sumlist(A,B):- head(A,B).
sumlist(A,B):- head(A,D),tail(A,C),sumlist(C,E),sum(D,E,B).
\end{lstlisting}

\centering
\begin{lstlisting}[caption=next]
next(A,B,C):- tail(A,D),head(D,C),head(A,B).
next(A,B,C):- tail(A,D),next(D,B,C).
\end{lstlisting}

\centering
\begin{lstlisting}[caption=twosucc]
twosucc(A) :- head(A,B), tail(A,C), head(C,D), increment(B,D).
twosucc(A) :- tail(A,C), twosucc(C).
\end{lstlisting}


\centering
\begin{lstlisting}[caption=rotateN]
rotateN(A,B,C) :- zero(A),eq(B,C).
rotateN(A,B,C) :- decrement(A,D),head(B,D),tail(B,F),append(F,D,E),rotateN(D,E,C).
\end{lstlisting}

\centering
\begin{lstlisting}[caption=inttobin]
inttobin(A,B):- empty(A), empty(B).
inttobin(A,B):- head(A,C),tail(A,D),bin(C,E),head(B,E),inttobin(D,F),tail(B,F).
\end{lstlisting}


\caption{Example solutions.}
\label{fig:sols}
\end{figure*}

\section{Experimental results}


Table \ref{tab:q1-600} shows the predictive accuracies of the systems when given a \textbf{30 minute} timeout.
Table \ref{tab:q1times} shows the termination times of the systems.
Table \ref{tab:num_progs_appendix} shows the number of programs generated.
Table \ref{tab:overhead_appendix} shows the overhead of discovering minimal unsatisfiable subprogram.
Figure \ref{fig:sols} shows example solutions.

\begin{table*}[ht!]
\centering
\small
\begin{tabular}{@{}l|ccccc@{}}
\textbf{Task} & 
\textbf{\name{}} & 
\textbf{\disco{}} & 
\textbf{\textsc{\popper{}}}& 
\textbf{\ale{}} & \textbf{\metagol{}}\\
\midrule
\emph{trains1} & 100 $\pm$ 0 & 100 $\pm$ 0 & 100 $\pm$ 0 & 100 $\pm$ 0 & 27 $\pm$ 0 \\
\emph{trains2} & 97 $\pm$ 1 & 98 $\pm$ 0.8 & 98 $\pm$ 0.8 & 100 $\pm$ 0 & 42 $\pm$ 12 \\
\emph{trains3} & 100 $\pm$ 0 & 100 $\pm$ 0 & 100 $\pm$ 0 & 100 $\pm$ 0 & 79 $\pm$ 0 \\
\emph{trains4} & 100 $\pm$ 0 & 100 $\pm$ 0 & 100 $\pm$ 0 & 100 $\pm$ 0 & 32 $\pm$ 0 \\
\midrule
\emph{zendo1} & 97 $\pm$ 0.7 & 97 $\pm$ 0.7 & 97 $\pm$ 0.7 & 90 $\pm$ 3 & 69 $\pm$ 8 \\
\emph{zendo2} & 100 $\pm$ 0.4 & 100 $\pm$ 0.4 & 100 $\pm$ 0.4 & 97 $\pm$ 1 & 50 $\pm$ 0 \\
\emph{zendo3} & 97 $\pm$ 1 & 95 $\pm$ 2 & 94 $\pm$ 2 & 97 $\pm$ 2 & 50 $\pm$ 0 \\
\emph{zendo4} & 95 $\pm$ 1 & 95 $\pm$ 1 & 95 $\pm$ 1 & 88 $\pm$ 1 & 50 $\pm$ 0 \\
\midrule
\emph{imdb1} & 100 $\pm$ 0 & 100 $\pm$ 0 & 100 $\pm$ 0 & 100 $\pm$ 0 & 16 $\pm$ 0 \\
\emph{imdb2} & 100 $\pm$ 0 & 100 $\pm$ 0 & 100 $\pm$ 0 & 50 $\pm$ 0 & 50 $\pm$ 0 \\
\emph{imdb3} & 100 $\pm$ 0 & 100 $\pm$ 0 & 100 $\pm$ 0 & 50 $\pm$ 0 & 50 $\pm$ 0 \\
\midrule
\emph{krk1} & 99 $\pm$ 0.3 & 99 $\pm$ 0.3 & 99 $\pm$ 0.3 & 97 $\pm$ 0.5 & 50 $\pm$ 0 \\
\emph{krk2} & 67 $\pm$ 6 & 59 $\pm$ 5 & 60 $\pm$ 5 & 95 $\pm$ 0.8 & 50 $\pm$ 0 \\
\emph{krk3} & 52 $\pm$ 0.5 & 52 $\pm$ 0.5 & 52 $\pm$ 0.5 & 92 $\pm$ 2 & 50 $\pm$ 0 \\
\midrule
\emph{md} & 100 $\pm$ 0 & 100 $\pm$ 0 & 100 $\pm$ 0 & 94 $\pm$ 0 & 11 $\pm$ 0 \\
\emph{buttons} & 98 $\pm$ 0.4 & 100 $\pm$ 0.1 & 100 $\pm$ 0 & 96 $\pm$ 0 & 19 $\pm$ 0 \\
\emph{rps} & 100 $\pm$ 0 & 100 $\pm$ 0 & 100 $\pm$ 0 & 100 $\pm$ 0 & 19 $\pm$ 0 \\
\emph{coins} & 100 $\pm$ 0 & 100 $\pm$ 0 & 100 $\pm$ 0 & 17 $\pm$ 0 & 17 $\pm$ 0 \\
\emph{buttons-g} & 96 $\pm$ 2 & 97 $\pm$ 2 & 99 $\pm$ 1.0 & 100 $\pm$ 0 & 50 $\pm$ 0 \\
\emph{coins-g} & 100 $\pm$ 0 & 100 $\pm$ 0 & 100 $\pm$ 0 & 100 $\pm$ 0 & 50 $\pm$ 0 \\
\emph{attrition} & 99 $\pm$ 0 & 99 $\pm$ 0 & 99 $\pm$ 0 & 93 $\pm$ 0 & 3 $\pm$ 0 \\
\emph{centipede} & 100 $\pm$ 0 & 100 $\pm$ 0 & 100 $\pm$ 0 & 100 $\pm$ 0 & 100 $\pm$ 0 \\
\midrule
\emph{dropk} & 100 $\pm$ 0 & 100 $\pm$ 0 & 100 $\pm$ 0 & 55 $\pm$ 5 & 50 $\pm$ 0 \\
\emph{droplast} & 100 $\pm$ 0 & 100 $\pm$ 0 & 100 $\pm$ 0 & 50 $\pm$ 0 & 50 $\pm$ 0 \\
\emph{evens} & 100 $\pm$ 0 & 100 $\pm$ 0 & 100 $\pm$ 0 & 50 $\pm$ 0.1 & 50 $\pm$ 0 \\
\emph{finddup} & 98 $\pm$ 0.5 & 98 $\pm$ 0.5 & 98 $\pm$ 0.5 & 51 $\pm$ 0.3 & 55 $\pm$ 5 \\
\emph{last} & 100 $\pm$ 0 & 100 $\pm$ 0 & 100 $\pm$ 0 & 50 $\pm$ 0.1 & 55 $\pm$ 5 \\
\emph{contains2} & 100 $\pm$ 0 & 100 $\pm$ 0 & 100 $\pm$ 0 & 54 $\pm$ 1.0 & 60 $\pm$ 7 \\
\emph{len} & 100 $\pm$ 0 & 100 $\pm$ 0 & 100 $\pm$ 0 & 50 $\pm$ 0 & 50 $\pm$ 0 \\
\emph{reverse} & 100 $\pm$ 0 & 95 $\pm$ 5 & 95 $\pm$ 5 & 50 $\pm$ 0 & 50 $\pm$ 0 \\
\emph{sorted} & 100 $\pm$ 0 & 100 $\pm$ 0 & 100 $\pm$ 0 & 73 $\pm$ 3 & 50 $\pm$ 0 \\
\emph{sumlist} & 100 $\pm$ 0 & 100 $\pm$ 0 & 100 $\pm$ 0 & 50 $\pm$ 0 & 100 $\pm$ 0 \\
\emph{next} & 100 $\pm$ 0 & 100 $\pm$ 0 & 100 $\pm$ 0 & 50 $\pm$ 0 & 50 $\pm$ 0 \\
\emph{rotateN} & 100 $\pm$ 0 & 95 $\pm$ 5 & 100 $\pm$ 0 & 50 $\pm$ 0 & 50 $\pm$ 0 \\
\emph{inttobin} & 99 $\pm$ 0.3 & 99 $\pm$ 0.2 & 100 $\pm$ 0.2 & 50 $\pm$ 0 & 50 $\pm$ 0 \\
\emph{chartointallodd} & 100 $\pm$ 0 & 100 $\pm$ 0 & 95 $\pm$ 5 & 52 $\pm$ 0.4 & 50 $\pm$ 0 \\
\emph{twosucc} & 72 $\pm$ 6 & 63 $\pm$ 4 & 59 $\pm$ 0.6 & 74 $\pm$ 6 & 50 $\pm$ 0 \\
\midrule
\emph{sql-01} & 100 $\pm$ 0 & 100 $\pm$ 0 & 100 $\pm$ 0 & 100 $\pm$ 0 & 0 $\pm$ 0 \\
\emph{sql-02} & 100 $\pm$ 0 & 100 $\pm$ 0 & 90 $\pm$ 10 & 100 $\pm$ 0 & 0 $\pm$ 0 \\
\emph{sql-03} & 100 $\pm$ 0 & 100 $\pm$ 0 & 80 $\pm$ 13 & 100 $\pm$ 0 & 0 $\pm$ 0 \\
\emph{sql-04} & 100 $\pm$ 0 & 85 $\pm$ 8 & 85 $\pm$ 11 & 100 $\pm$ 0 & 0 $\pm$ 0 \\
\emph{sql-05} & 100 $\pm$ 0 & 100 $\pm$ 0 & 100 $\pm$ 0 & 100 $\pm$ 0 & 0 $\pm$ 0 \\
\emph{sql-06} & 100 $\pm$ 0 & 100 $\pm$ 0 & 100 $\pm$ 0 & 100 $\pm$ 0 & 0 $\pm$ 0 \\
\emph{sql-07} & 100 $\pm$ 0 & 100 $\pm$ 0 & 100 $\pm$ 0 & 100 $\pm$ 0 & 0 $\pm$ 0 \\
\emph{sql-08} & 100 $\pm$ 0 & 90 $\pm$ 10 & 40 $\pm$ 16 & 100 $\pm$ 0 & 0 $\pm$ 0 \\
\emph{sql-09} & 100 $\pm$ 0 & 100 $\pm$ 0 & 100 $\pm$ 0 & 100 $\pm$ 0 & 0 $\pm$ 0 \\
\emph{sql-10} & 100 $\pm$ 0 & 100 $\pm$ 0 & 100 $\pm$ 0 & 100 $\pm$ 0 & 0 $\pm$ 0 \\
\emph{sql-11} & 100 $\pm$ 0 & 50 $\pm$ 17 & 30 $\pm$ 15 & 100 $\pm$ 0 & 0 $\pm$ 0 \\
\emph{sql-12} & 100 $\pm$ 0 & 100 $\pm$ 0 & 100 $\pm$ 0 & 100 $\pm$ 0 & 0 $\pm$ 0 \\
\emph{sql-13} & 100 $\pm$ 0 & 100 $\pm$ 0 & 100 $\pm$ 0 & 100 $\pm$ 0 & 100 $\pm$ 0 \\
\emph{sql-14} & 100 $\pm$ 0 & 100 $\pm$ 0 & 100 $\pm$ 0 & 100 $\pm$ 0 & 0 $\pm$ 0 \\
\emph{sql-15} & 100 $\pm$ 0 & 100 $\pm$ 0 & 100 $\pm$ 0 & 100 $\pm$ 0 & 100 $\pm$ 0 \\

\end{tabular}
\caption{
Predictive accuracies with a 30 minute learning timeout.
We round accuracies to integer values. The error is standard error.
}
\label{tab:q1-600}
\end{table*}


\begin{table*}[ht!]
\small
\centering
\begin{tabular}{@{}l|ccccc@{}}
\textbf{Task} & 
\textbf{\name{}} & 
\textbf{\disco{}} & 
\textbf{\popper{}} & 
\textbf{\ale{}} & 
\textbf{\metagol{}}\\
\midrule
\emph{trains1} & 6 $\pm$ 0.1 & 6 $\pm$ 0.2 & 6 $\pm$ 0.1 & \textbf{2 $\pm$ 0.2}  & \emph{timeout} \\
\emph{trains2} & 6 $\pm$ 0.2 & 7 $\pm$ 0.4 & 6 $\pm$ 0.2 & \textbf{1 $\pm$ 0.1}  & 1571 $\pm$ 117 \\
\emph{trains3} & 21 $\pm$ 0.4 & 22 $\pm$ 0.9 & 21 $\pm$ 0.4 & \textbf{5 $\pm$ 0.5}  & \emph{timeout} \\
\emph{trains4} & \textbf{18 $\pm$ 0.3}  & 20 $\pm$ 0.8 & \textbf{18 $\pm$ 0.3}  & 20 $\pm$ 2 & \emph{timeout} \\
\midrule
\emph{zendo1} & 4 $\pm$ 0.4 & 4 $\pm$ 0.2 & 4 $\pm$ 0.3 & \textbf{3 $\pm$ 1.0}  & 1082 $\pm$ 293 \\
\emph{zendo2} & 61 $\pm$ 6 & 68 $\pm$ 9 & 61 $\pm$ 7 & \textbf{1 $\pm$ 0.2}  & \emph{timeout} \\
\emph{zendo3} & 46 $\pm$ 4 & 51 $\pm$ 8 & 48 $\pm$ 6 & \textbf{1 $\pm$ 0.2}  & \emph{timeout} \\
\emph{zendo4} & 32 $\pm$ 6 & 34 $\pm$ 7 & 32 $\pm$ 7 & \textbf{1 $\pm$ 0.2}  & \emph{timeout} \\
\midrule
\emph{imdb1} & \textbf{4 $\pm$ 0.1}  & \textbf{4 $\pm$ 0.1}  & \textbf{4 $\pm$ 0}  & 101 $\pm$ 29 & \emph{timeout} \\
\emph{imdb2} & \textbf{4 $\pm$ 0.1}  & \textbf{4 $\pm$ 0.1}  & \textbf{4 $\pm$ 0.1}  & \emph{timeout} & \emph{timeout} \\
\emph{imdb3} & \textbf{435 $\pm$ 34}  & 741 $\pm$ 108 & 571 $\pm$ 37 & \emph{timeout} & \emph{timeout} \\
\midrule
\emph{krk1} & 38 $\pm$ 5 & 53 $\pm$ 10 & 41 $\pm$ 6 & \textbf{0.4 $\pm$ 0}  & 87 $\pm$ 0.6 \\
\emph{krk2} & \emph{timeout} & \emph{timeout} & \emph{timeout} & \textbf{8 $\pm$ 2}  & 96 $\pm$ 0.5 \\
\emph{krk3} & \emph{timeout} & \emph{timeout} & \emph{timeout} & \textbf{22 $\pm$ 7}  & 88 $\pm$ 0.6 \\
\midrule
\emph{md} & 53 $\pm$ 2 & 253 $\pm$ 26 & 270 $\pm$ 25 & \textbf{4 $\pm$ 0.1}  & \emph{timeout} \\
\emph{buttons} & \textbf{50 $\pm$ 4}  & 891 $\pm$ 185 & 743 $\pm$ 144 & 64 $\pm$ 0.3 & \emph{timeout} \\
\emph{rps} & 162 $\pm$ 2 & 177 $\pm$ 13 & 185 $\pm$ 2 & 10 $\pm$ 0.1 & \textbf{0.1 $\pm$ 0}  \\
\emph{coins} & 952 $\pm$ 63 & 698 $\pm$ 25 & 1028 $\pm$ 71 & \emph{timeout} & \textbf{0.2 $\pm$ 0}  \\
\emph{buttons-g} & 12 $\pm$ 0.2 & 19 $\pm$ 1 & 21 $\pm$ 0.5 & 47 $\pm$ 0.5 & \textbf{0.1 $\pm$ 0.1}  \\
\emph{coins-g} & 6 $\pm$ 0.1 & \emph{timeout} & \emph{timeout} & 8 $\pm$ 0.2 & \textbf{0.1 $\pm$ 0}  \\
\emph{attrition} & 290 $\pm$ 24 & 685 $\pm$ 42 & 740 $\pm$ 11 & 0.9 $\pm$ 0 & \textbf{0.1 $\pm$ 0}  \\
\emph{centipede} & 129 $\pm$ 11 & 136 $\pm$ 7 & 147 $\pm$ 2 & 7 $\pm$ 0.1 & \textbf{0.1 $\pm$ 0}  \\
\midrule
\emph{dropk} & 7 $\pm$ 0.8 & 10 $\pm$ 1 & 10 $\pm$ 1 & 903 $\pm$ 299 & \textbf{0.1 $\pm$ 0}  \\
\emph{droplast} & \textbf{19 $\pm$ 1}  & 25 $\pm$ 2 & 27 $\pm$ 2 & 1011 $\pm$ 92 & \emph{timeout} \\
\emph{evens} & 38 $\pm$ 3 & 77 $\pm$ 8 & 58 $\pm$ 7 & \textbf{2 $\pm$ 0}  & \emph{timeout} \\
\emph{finddup} & \textbf{74 $\pm$ 11}  & 78 $\pm$ 12 & 133 $\pm$ 12 & 725 $\pm$ 293 & 1621 $\pm$ 180 \\
\emph{last} & 11 $\pm$ 0.9 & 12 $\pm$ 1 & 22 $\pm$ 2 & \textbf{2 $\pm$ 0.1}  & 1701 $\pm$ 99 \\
\emph{contains2} & 268 $\pm$ 12 & 255 $\pm$ 11 & 306 $\pm$ 11 & \textbf{56 $\pm$ 2}  & 1452 $\pm$ 232 \\
\emph{len} & 24 $\pm$ 5 & 47 $\pm$ 8 & 57 $\pm$ 6 & \textbf{2 $\pm$ 0.2}  & \emph{timeout} \\
\emph{reverse} & \textbf{203 $\pm$ 19}  & 844 $\pm$ 134 & 1053 $\pm$ 149 & 673 $\pm$ 98 & \emph{timeout} \\
\emph{sorted} & 216 $\pm$ 26 & 239 $\pm$ 26 & 253 $\pm$ 25 & \textbf{1.0 $\pm$ 0.1}  & \emph{timeout} \\
\emph{sumlist} & 25 $\pm$ 0.8 & 263 $\pm$ 39 & 276 $\pm$ 49 & \textbf{0.3 $\pm$ 0}  & \textbf{0.3 $\pm$ 0}  \\
\emph{next} & 44 $\pm$ 7 & 57 $\pm$ 8 & 59 $\pm$ 10 & 0.3 $\pm$ 0 & \textbf{0.1 $\pm$ 0}  \\
\emph{rotateN} & 292 $\pm$ 49 & 819 $\pm$ 156 & 437 $\pm$ 43 & 0.3 $\pm$ 0 & \textbf{0.1 $\pm$ 0}  \\
\emph{inttobin} & 65 $\pm$ 8 & 59 $\pm$ 8 & 83 $\pm$ 12 & \textbf{0.6 $\pm$ 0}  & \emph{timeout} \\
\emph{chartointallodd} & 104 $\pm$ 5 & 429 $\pm$ 43 & 638 $\pm$ 149 & \textbf{0.6 $\pm$ 0}  & \emph{timeout} \\
\emph{twosucc} & 1764 $\pm$ 23 & 1765 $\pm$ 37 & \emph{timeout} & \textbf{2 $\pm$ 0.4}  & \emph{timeout} \\
\midrule
\emph{sql-01} & 6 $\pm$ 0.1 & 7 $\pm$ 1 & 14 $\pm$ 6 & \textbf{0.3 $\pm$ 0}  & 79 $\pm$ 0.8 \\
\emph{sql-02} & 29 $\pm$ 1 & 1005 $\pm$ 136 & 936 $\pm$ 136 & \textbf{0.3 $\pm$ 0}  & 130 $\pm$ 0.9 \\
\emph{sql-03} & 24 $\pm$ 2 & 700 $\pm$ 129 & 1126 $\pm$ 138 & \textbf{0.3 $\pm$ 0}  & 133 $\pm$ 0.9 \\
\emph{sql-04} & 71 $\pm$ 4 & \emph{timeout} & \emph{timeout} & \textbf{0.3 $\pm$ 0}  & \emph{timeout} \\
\emph{sql-05} & 5 $\pm$ 0.1 & 6 $\pm$ 0.4 & 8 $\pm$ 0.1 & \textbf{0.3 $\pm$ 0}  & 80 $\pm$ 0.5 \\
\emph{sql-06} & 14 $\pm$ 0.8 & \emph{timeout} & \emph{timeout} & 0.3 $\pm$ 0 & \textbf{0.1 $\pm$ 0}  \\
\emph{sql-07} & 6 $\pm$ 0.1 & 7 $\pm$ 0.8 & 9 $\pm$ 0.4 & \textbf{0.3 $\pm$ 0}  & 132 $\pm$ 1 \\
\emph{sql-08} & 8 $\pm$ 0.2 & 941 $\pm$ 191 & 1527 $\pm$ 142 & 0.3 $\pm$ 0 & \textbf{0.1 $\pm$ 0}  \\
\emph{sql-09} & 4 $\pm$ 0.2 & 4 $\pm$ 0.2 & 5 $\pm$ 0.1 & 0.3 $\pm$ 0 & \textbf{0.1 $\pm$ 0}  \\
\emph{sql-10} & 6 $\pm$ 0.1 & 12 $\pm$ 0.7 & 11 $\pm$ 0.8 & 0.3 $\pm$ 0 & \textbf{0.1 $\pm$ 0}  \\
\emph{sql-11} & 8 $\pm$ 0.2 & 1237 $\pm$ 203 & 1615 $\pm$ 125 & \textbf{0.3 $\pm$ 0}  & 89 $\pm$ 0.6 \\
\emph{sql-12} & 5 $\pm$ 0.2 & 40 $\pm$ 8 & 41 $\pm$ 11 & \textbf{0.3 $\pm$ 0}  & 85 $\pm$ 0.5 \\
\emph{sql-13} & 5 $\pm$ 0.1 & 4 $\pm$ 0.2 & 5 $\pm$ 0.1 & 0.3 $\pm$ 0 & \textbf{0.1 $\pm$ 0}  \\
\emph{sql-14} & 5 $\pm$ 0.1 & 4 $\pm$ 0.2 & 5 $\pm$ 0.1 & 0.3 $\pm$ 0 & \textbf{0.1 $\pm$ 0}  \\
\emph{sql-15} & 5 $\pm$ 0.2 & 5 $\pm$ 0.3 & 5 $\pm$ 0.2 & 0.3 $\pm$ 0 & \textbf{0.1 $\pm$ 0}  \\
\end{tabular}
\caption{
Learning times given a 30 minute timeout.
A \emph{timeout} entry means that the system did not terminate in the given time.
We round times over one second to the nearest second.
The error is standard error.
}
\label{tab:q1times}
\end{table*}

\begin{table}[t]
\centering
\begin{tabular}{@{}l|ccc@{}}
    \textbf{Task} & \textbf{\popper{}} & \textbf{\name{}} & \textbf{Change}\\
    \midrule
\emph{trains1} & 686 $\pm$ 0 & 591 $\pm$ 0 & \textbf{-13\%} \\
\emph{trains2} & 657 $\pm$ 8 & 666 $\pm$ 8 & +1\% \\
\emph{trains3} & 2537 $\pm$ 1 & 2537 $\pm$ 1 & 0\% \\
\emph{trains4} & 2712 $\pm$ 0 & 2712 $\pm$ 0 & 0\% \\
\midrule
\emph{zendo1} & 869 $\pm$ 308 & 1032 $\pm$ 466 & +18\% \\
\emph{zendo2} & 3434 $\pm$ 405 & 3388 $\pm$ 455 & \textbf{-1\%} \\
\emph{zendo3} & 2982 $\pm$ 279 & 3150 $\pm$ 263 & +5\% \\
\emph{zendo4} & 2340 $\pm$ 400 & 2365 $\pm$ 414 & +1\% \\
\midrule
\emph{imdb1} & 4 $\pm$ 0 & 3 $\pm$ 0 & \textbf{-25\%} \\
\emph{imdb2} & 41 $\pm$ 0 & 45 $\pm$ 0 & +9\% \\
\emph{imdb3} & 521 $\pm$ 0.9 & 478 $\pm$ 0.4 & \textbf{-8\%} \\
\midrule
\emph{krk1} & 427 $\pm$ 69 & 375 $\pm$ 59 & \textbf{-12\%} \\
\emph{krk2} & 5414 $\pm$ 1431 & 9246 $\pm$ 404 & +70\% \\
\emph{krk3} & 2858 $\pm$ 152 & 2892 $\pm$ 348 & +1\% \\
\midrule
\emph{md} & 2337 $\pm$ 51 & 652 $\pm$ 8 & \textbf{-72\%} \\
\emph{buttons} & 4539 $\pm$ 28 & 1274 $\pm$ 18 & \textbf{-71\%} \\
\emph{rps} & 11259 $\pm$ 5 & 8771 $\pm$ 4 & \textbf{-22\%} \\
\emph{coins} & 62239 $\pm$ 1644 & 61518 $\pm$ 1672 & \textbf{-1\%} \\
\emph{buttons-g} & 1298 $\pm$ 0.7 & 576 $\pm$ 5 & \textbf{-55\%} \\
\emph{coins-g} & 58378 $\pm$ 0 & 286 $\pm$ 0.2 & \textbf{-99\%} \\
\emph{attrition} & 159070 $\pm$ 1280 & 56948 $\pm$ 465 & \textbf{-64\%} \\
\emph{centipede} & 2307 $\pm$ 0 & 1925 $\pm$ 154 & \textbf{-16\%} \\
\midrule
\emph{dropk} & 235 $\pm$ 24 & 161 $\pm$ 15 & \textbf{-31\%} \\
\emph{droplast} & 95 $\pm$ 4 & 93 $\pm$ 2 & \textbf{-2\%} \\
\emph{evens} & 301 $\pm$ 3 & 300 $\pm$ 3 & \textbf{0\%} \\
\emph{finddup} & 2167 $\pm$ 79 & 1444 $\pm$ 63 & \textbf{-33\%} \\
\emph{last} & 388 $\pm$ 44 & 165 $\pm$ 29 & \textbf{-57\%} \\
\emph{contains2} & 1382 $\pm$ 22 & 1340 $\pm$ 21 & \textbf{-3\%} \\
\emph{len} & 864 $\pm$ 87 & 337 $\pm$ 70 & \textbf{-60\%} \\
\emph{reverse} & 2304 $\pm$ 141 & 474 $\pm$ 63 & \textbf{-79\%} \\
\emph{sorted} & 942 $\pm$ 50 & 845 $\pm$ 57 & \textbf{-10\%} \\
\emph{sumlist} & 1391 $\pm$ 3 & 85 $\pm$ 3 & \textbf{-93\%} \\
\emph{next} & 1588 $\pm$ 126 & 365 $\pm$ 61 & \textbf{-77\%} \\
\emph{rotateN} & 8164 $\pm$ 1574 & 1164 $\pm$ 167 & \textbf{-85\%} \\
\emph{inttobin} & 4788 $\pm$ 342 & 1430 $\pm$ 135 & \textbf{-70\%} \\
\emph{chartointallodd} & 787 $\pm$ 21 & 354 $\pm$ 24 & \textbf{-55\%} \\
\emph{twosucc} & 1311 $\pm$ 34 & 1284 $\pm$ 25 & \textbf{-2\%} \\
\midrule
\emph{sql-01} & 297 $\pm$ 0 & 15 $\pm$ 0.3 & \textbf{-94\%} \\
\emph{sql-02} & 28751 $\pm$ 0 & 49 $\pm$ 0.2 & \textbf{-99\%} \\
\emph{sql-03} & 25421 $\pm$ 0 & 55 $\pm$ 0 & \textbf{-99\%} \\
\emph{sql-04} & 23993 $\pm$ 0 & 44 $\pm$ 0.2 & \textbf{-99\%} \\
\emph{sql-05} & 272 $\pm$ 0 & 7 $\pm$ 0.3 & \textbf{-97\%} \\
\emph{sql-06} & 444 $\pm$ 0 & 48 $\pm$ 0 & \textbf{-89\%} \\
\emph{sql-07} & 578 $\pm$ 0 & 15 $\pm$ 0 & \textbf{-97\%} \\
\emph{sql-08} & 116043 $\pm$ 0 & 33 $\pm$ 0.4 & \textbf{-99\%} \\
\emph{sql-09} & 1 $\pm$ 0 & 1 $\pm$ 0 & 0\% \\
\emph{sql-10} & 199 $\pm$ 0 & 36 $\pm$ 0 & \textbf{-81\%} \\
\emph{sql-11} & 4093 $\pm$ 0 & 40 $\pm$ 0.3 & \textbf{-99\%} \\
\emph{sql-12} & 321 $\pm$ 0 & 24 $\pm$ 0 & \textbf{-92\%} \\
\emph{sql-13} & 5 $\pm$ 0 & 5 $\pm$ 0 & 0\% \\
\emph{sql-14} & 2 $\pm$ 0 & 2 $\pm$ 0 & 0\% \\
\emph{sql-15} & 140 $\pm$ 0 & 6 $\pm$ 0 & \textbf{-95\%} \\
\end{tabular}
\caption{
Number of programs generated for \popper{} and \name{}.
Error is standard error. 
}
\label{tab:num_progs_appendix}
\end{table}

\begin{table}[ht]
\centering
\begin{tabular}{@{}l|ccc@{}}
    \textbf{Task} & \textbf{Learning time} & \textbf{Explain time} & \textbf{Ratio}\\
    \midrule
\emph{trains1} & 4 $\pm$ 0.2 & 0 $\pm$ 0 & 0\% \\
\emph{trains2} & 4 $\pm$ 0.3 & 0 $\pm$ 0 & 0\% \\
\emph{trains3} & 24 $\pm$ 0.1 & 0 $\pm$ 0 & 0\% \\
\emph{trains4} & 20 $\pm$ 0.1 & 0 $\pm$ 0 & 0\% \\
\midrule
\emph{zendo1} & 11 $\pm$ 5 & 0.1 $\pm$ 0 & 0\% \\
\emph{zendo2} & 53 $\pm$ 10 & 0.2 $\pm$ 0 & 0\% \\
\emph{zendo3} & 49 $\pm$ 6 & 0.2 $\pm$ 0 & 0\% \\
\emph{zendo4} & 31 $\pm$ 5 & 0.1 $\pm$ 0 & 0\% \\
\midrule
\emph{imdb1} & 0.9 $\pm$ 0.1 & 0 $\pm$ 0 & 0\% \\
\emph{imdb2} & 2 $\pm$ 0.1 & 0 $\pm$ 0 & 0\% \\
\emph{imdb3} & 658 $\pm$ 49 & 9 $\pm$ 2 & 1\% \\
\midrule
\emph{krk1} & 34 $\pm$ 5 & 0.7 $\pm$ 0.1 & 2\% \\
\midrule
\emph{md} & 10 $\pm$ 0.5 & 0.9 $\pm$ 0.1 & 9\% \\
\emph{buttons} & 14 $\pm$ 1 & 0.6 $\pm$ 0 & 4\% \\
\emph{rps} & 110 $\pm$ 6 & 3 $\pm$ 0.2 & 2\% \\
\emph{coins} & 517 $\pm$ 23 & 1 $\pm$ 0.1 & 0\% \\
\emph{buttons-g} & 3 $\pm$ 0.1 & 0.1 $\pm$ 0 & 3\% \\
\emph{coins-g} & 3 $\pm$ 0.1 & 0.2 $\pm$ 0 & 6\% \\
\emph{attrition} & 195 $\pm$ 11 & 8 $\pm$ 0.2 & 4\% \\
\emph{centipede} & 27 $\pm$ 2 & 3 $\pm$ 0.3 & 11\% \\
\midrule
\emph{dropk} & 6 $\pm$ 2 & 0.1 $\pm$ 0 & 1\% \\
\emph{droplast} & 10 $\pm$ 0.8 & 0 $\pm$ 0 & 0\% \\
\emph{evens} & 5 $\pm$ 0.1 & 0 $\pm$ 0 & 0\% \\
\emph{finddup} & 7 $\pm$ 0.2 & 0.3 $\pm$ 0 & 4\% \\
\emph{last} & 3 $\pm$ 0.1 & 0.1 $\pm$ 0 & 3\% \\
\emph{contains2} & 13 $\pm$ 0.4 & 0 $\pm$ 0 & 0\% \\
\emph{len} & 4 $\pm$ 0.2 & 0.1 $\pm$ 0 & 2\% \\
\emph{reverse} & 94 $\pm$ 15 & 4 $\pm$ 0.7 & 4\% \\
\emph{sorted} & 9 $\pm$ 0.5 & 0.1 $\pm$ 0 & 1\% \\
\emph{sumlist} & 3 $\pm$ 0 & 0 $\pm$ 0 & 0\% \\
\emph{next} & 35 $\pm$ 8 & 0.1 $\pm$ 0 & 0\% \\
\emph{rotateN} & 322 $\pm$ 47 & 21 $\pm$ 4 & 6\% \\
\emph{inttobin} & 45 $\pm$ 4 & 28 $\pm$ 2 & 62\% \\
\emph{chartointallodd} & 5 $\pm$ 0.2 & 0.1 $\pm$ 0 & 1\% \\
\emph{twosucc} & 14 $\pm$ 0.9 & 0.1 $\pm$ 0 & 0\% \\
\midrule
\emph{sql-01} & 0.7 $\pm$ 0 & 0.3 $\pm$ 0 & 42\% \\
\emph{sql-02} & 3 $\pm$ 0 & 2 $\pm$ 0 & 66\% \\
\emph{sql-03} & 3 $\pm$ 0 & 2 $\pm$ 0 & 66\% \\
\emph{sql-04} & 0.7 $\pm$ 0 & 0.2 $\pm$ 0 & 28\% \\
\emph{sql-05} & 0.3 $\pm$ 0 & 0.1 $\pm$ 0 & 33\% \\
\emph{sql-06} & 0.3 $\pm$ 0 & 0.1 $\pm$ 0 & 33\% \\
\emph{sql-07} & 0.7 $\pm$ 0 & 0.3 $\pm$ 0 & 42\% \\
\emph{sql-08} & 0.9 $\pm$ 0 & 0.4 $\pm$ 0 & 44\% \\
\emph{sql-09} & 0.2 $\pm$ 0 & 0 $\pm$ 0 & 0\% \\
\emph{sql-10} & 0.3 $\pm$ 0 & 0.1 $\pm$ 0 & 33\% \\
\emph{sql-11} & 1 $\pm$ 0 & 0.4 $\pm$ 0 & 40\% \\
\emph{sql-12} & 0.6 $\pm$ 0 & 0.1 $\pm$ 0 & 16\% \\
\emph{sql-13} & 0.2 $\pm$ 0 & 0 $\pm$ 0 & 0\% \\
\emph{sql-14} & 0.3 $\pm$ 0 & 0 $\pm$ 0 & 0\% \\
\emph{sql-15} & 0.2 $\pm$ 0 & 0 $\pm$ 0 & 0\% \\
\end{tabular}
\caption{
Overhead (in seconds) of discovering minimal unsatisfiable subprograms.
We round times over one second to the nearest second.
Error is standard error.
}
\label{tab:overhead_appendix}
\end{table}

 \end{appendices}

\end{document}